\documentclass{article} 
\usepackage{iclr2027_conference,times}

\usepackage{amsmath,amsfonts,bm}

\def\eqref#1{equation~\ref{#1}}

\def\1{\bm{1}}

\DeclareMathAlphabet{\mathsfit}{\encodingdefault}{\sfdefault}{m}{sl}
\SetMathAlphabet{\mathsfit}{bold}{\encodingdefault}{\sfdefault}{bx}{n}

\usepackage{url}
\usepackage[table]{xcolor}
\usepackage{booktabs}
\usepackage{tabularx}
\usepackage{graphicx}
\usepackage{wrapfig}
\usepackage{enumitem}
\usepackage{float}
\usepackage[most]{tcolorbox}

\newtcolorbox{promptbox}[1]{
  enhanced,
  breakable,
  colback=black!1,
  colframe=black!40,
  colbacktitle=black!7,
  coltitle=black,
  fonttitle=\bfseries,
  title={#1},
  boxrule=0.5pt,
  arc=1.2mm,
  left=2mm,
  right=2mm,
  top=1.5mm,
  bottom=1.5mm,
  before skip=7pt,
  after skip=7pt
}
\usepackage{hyperref}

\newcommand{\headlogo}[2]{\smash{\includegraphics[height=#1]{#2}}}
\renewcommand{\headrulewidth}{0.7pt}
\fancypagestyle{preprintfirst}{%
  \fancyhf{}%
  \fancyhead[L]{\headlogo{23pt}{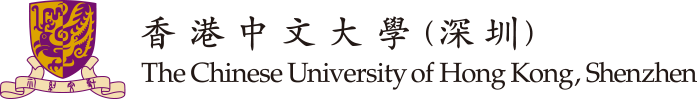}}%
  \fancyhead[R]{\headlogo{15pt}{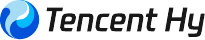}}%
  \fancyfoot[C]{\thepage}%
}
\fancypagestyle{preprintrest}{%
  \fancyhf{}%
  \fancyfoot[C]{\thepage}%
}

\title{AURAL: Adaptive Latent Reasoning with Joint Chunk for Speech Language Models}

\author{
  \textbf{Yuxiang Wang}$^{1,2,*}$,
  \textbf{Kunyu Feng}$^{1}$,
  \textbf{Yuancheng Wang}$^{1}$,
  \textbf{Zihang Liu}$^{4}$,
  \textbf{Shengbo Cai}$^{2,4}$\\
  \textbf{Qinke Ni}$^{1}$,
  \textbf{Wan Lin}$^{1}$,
  \textbf{Tao Feng}$^{5}$,
  \textbf{Yingda shen}$^{1}$,
  \textbf{Ming-Hao Hsu}$^{1}$,
  \textbf{Zhixian Zhao}$^{2}$\\
  \textbf{Liqiang Zhang}$^{2}$,
  \textbf{Teddy Sun}$^{2}$,
  \textbf{Steve Yves}$^{2}$,
  \textbf{Zhizheng Wu}$^{1,3,\dagger}$ \\
  $^{1}$The Chinese University of Hong Kong, Shenzhen\quad $^{2}$Tencent Hunyuan\\
  $^{3}$Amphion Technology Co., Ltd.\quad
  $^{4}$Tsinghua University \\
  $^{5}$The Hong Kong University of Science and Technology
}

\iclrfinalcopy 
\begin{document}

\maketitle
\pagestyle{preprintrest}
\thispagestyle{preprintfirst}

\begingroup
\renewcommand{\thefootnote}{\fnsymbol{footnote}}
\footnotetext[1]{Work done during internship at Tencent Hunyuan.}
\footnotetext[2]{Corresponding author.}
\endgroup

\begin{abstract}
Model intelligence and fast response jointly shape the quality of interaction with speech language models, yet remain difficult to achieve together. Explicit chain-of-thought (CoT) improves reasoning and audio understanding, but generating intermediate reasoning tokens delays responses. Describing fine-grained acoustic cues further lengthens CoT and increases latency. Latent reasoning can reduce this overhead, yet existing methods often trail CoT and remain limited by single-path supervision and reasoning budgets that do not adapt to problem difficulty. We introduce \textbf{AURAL}\footnote{Gitub: \url{https://github.com/JensenYX/AURAL}}, which models a distribution over multiple plausible reasoning continuations in latent space and jointly predicts chunks of future states to reduce sequential forward passes and reasoning latency. To provide initial supervision for latent reasoning, we construct AuralReason-683K: 683K bilingual speech utterances (about 1,000 hours) with concise CoT for emotion recognition, empathetic dialogue, and general reasoning. AURAL-RL then explores beyond these traces, rewarding concise reasoning that yields high-quality answers and adapting reasoning effort to each problem. Across two backbones, AURAL-RL achieves performance comparable to CoT-RL, with larger gains over the respective supervised checkpoints on most metrics. Analysis further shows that harder questions elicit more latent reasoning steps. On Qwen2.5-Omni, it reduces time to the first answer token by 11.8×, from 1.22 to 0.10 s, versus 0.05 s for direct answering.
\end{abstract}

\vspace{-2mm}
\section{Introduction}
\label{sec:introduction}
\vspace{-2mm}

Interaction is one of the defining capabilities of a speech language model (SLM). Users will not keep talking to a model that lacks the intelligence to understand their needs and reason well enough to help. Nor will they wait through a long pause for every thoughtful answer. The ideal resembles the voice assistant imagined in \emph{Her}\footnote{\emph{Her} is a science fiction film released in 2013.}, responsive enough to feel present and perceptive enough to understand what the user means. Although current SLMs remain far from this vision, improving both intelligence and response speed is essential to closing the gap.

Explicit chain of thought (CoT) is a natural way to improve intelligence in LLMs, and in SLMs its benefits extend further to audio understanding \citep{tian2025stepaudio}. Yet CoT remains rare in interactive models because every intermediate reasoning token is generated before the answer and is therefore paid for in response latency. For speech, CoT introduces a second cost. Acoustic cues such as timbre, prosody, emotion, speaker traits, and background sounds vary continuously. Changes in these cues can alter how an utterance is understood, but text often reduces them to simple labels. The label ``upset'' can describe the tone of ``I am fine'', but it does not capture the strength and mixture of hesitation, tension, and sadness in the voice. Describing these details in CoT requires more words, which add tokens and further increase latency. Recent work finds that even when CoT is used, SLMs may recognize paralinguistic cues without using them in their responses \citep{wang2026voxsafebench,wang2026parabridge}. CoT therefore offers a natural path to more capable SLMs, yet its serial generation slows responses and its discrete form forces a trade between acoustic precision and speed.

Continuous hidden states can retain acoustic detail without spelling it out token by token. Latent reasoning performs intermediate computation in these states, reducing the number of serial reasoning steps before an answer. Each state can also represent information that does not map cleanly to a single vocabulary token, giving the model a broader reasoning space than explicit CoT. However, latent reasoning remains rare in SLMs. FLAIR fills the silence slots that occur while the user is speaking with latent thoughts, so speech length determines the number of reasoning steps \citep{wu2026silentthought}. CoAT and LatentOmni instead use a latent budget fixed in advance \citep{han2026coat,dai2026latentomni}. Although these designs limit response latency, they often lag behind explicit CoT in reasoning quality. Their reasoning budgets do not adapt to problem difficulty, and supervision typically provides a single target for each latent step, encouraging one continuation rather than preserving plausible alternatives. In addition, explicit CoT remains useful as initial supervision for latent reasoning, yet large speech corpora with reliable traces remain scarce.

We therefore propose \textbf{AURAL} (\textbf{A}daptive \textbf{U}nspoken \textbf{R}easoning over \textbf{A}lternative \textbf{L}atents), a latent reasoning framework for SLMs that retains
the benefits of CoT at much lower response latency (Figure 1).
AURAL models multiple plausible reasoning continuations in latent space
with a Gaussian mixture model and jointly predicts chunks of future
states to reduce sequential forward passes.
To provide initial supervision, we construct AuralReason-683K, a carefully
filtered bilingual collection of 683K utterances (about 1,000 hours) with
concise CoT traces for speech emotion recognition, empathetic dialogue,
and general reasoning.
AURAL-SFT pools consecutive CoT embeddings into latent targets.
AURAL-RL then explores beyond these traces, rewarding concise reasoning
that yields high-quality answers and learning a problem-dependent
stopping policy.
Across two SLM backbones, AURAL-RL achieves performance comparable to
CoT-RL, with larger gains over the respective supervised checkpoints
on most metrics.
Analysis on GPQA and VoiceBench BBH further shows that harder questions
elicit more latent reasoning steps.
With the default setting on Qwen2.5-Omni, it reduces time to the first
answer token from 1.22 s for CoT-RL to 0.10 s (an $11.8\times$ speedup),
compared with 0.05 s for direct answering.
Our contributions are:

\begin{itemize}[leftmargin=*,itemsep=1pt,topsep=2pt]    

\item We introduce AURAL, which models a distribution over plausible reasoning continuations in latent space and jointly predicts future-state chunks to reduce forward passes and reasoning latency.

    \item We construct AuralReason-683K, a bilingual corpus of 683K speech
    utterances (about 1,000 hours) with concise CoT for emotion recognition,
    empathetic dialogue, and general reasoning.

    \item We develop AURAL-RL to explore beyond supervised traces and adapt
    reasoning effort to each problem. It achieves performance comparable
    to CoT-RL with an $11.8\times$ speedup in time to the first answer token
    on Qwen2.5-Omni.
\end{itemize}
\vspace{-2mm}
\section{Related Work}
\label{sec:related_work}
\vspace{-2mm}

Latent reasoning moves intermediate computation from verbalized CoT into continuous states. Coconut feeds predicted hidden states back into the model \citep{hao2025coconut}, while CODI adds endpoint self-distillation \citep{shen2025codi}. Because neither method provides targets for intermediate latent states, the latent chain cannot be trained efficiently with teacher forcing and must instead be unrolled serially. Their preset latent length also prevents depth from adapting to problem difficulty. CCoT supplies teacher-forced targets, which improves supervision but still generates them autoregressively \citep{cheng2024ccot}. PCCoT replaces slow serial update with parallel Jacobi refinement at the cost of several refinement rounds \citep{wu2025pccot}. Latent-SFT obtains fine-grained supervision targets from a separately trained encoder \citep{deng2025latentsft}, while CoLaR \citep{colar2025} and C-MTP \citep{yerram2026cmtp} greatly simplify target construction by pooling grouped CoT embeddings but providing coarser supervision. Beyond supervised target design, Latent-GRPO shows that latent policy optimization can leave the valid manifold or average incompatible modes \citep{deng2026latentgrpo}. Latent Thought Flow instead learns stochastic trajectories whose length depends on answer quality and computation cost \citep{zou2026ltf}. These findings motivate latent reasoning that preserves multiple valid continuations and uses more computation only when a problem requires it.

Latent reasoning in SLMs remains sparse. FLAIR is the closest full-duplex precedent \citep{wu2026silentthought}. It replaces idle listening slots with recursively fed soft vocabulary embeddings learned from a global expert, enabling causal thought while the user speaks. Its reasoning depth is nevertheless determined by the incoming speech schedule rather than problem difficulty. CoAT inserts a latent workspace supervised by hand-selected audio experts \citep{han2026coat}. Because both the workspace and its supervisory dimensions are prescribed, it learns an engineered acoustic representation rather than discovering an autonomous reasoning trajectory. HyPeR invokes pause states at acoustically ambiguous points \citep{wang2026hyper}. This adds perception compute where confidence is low, but does not optimize reasoning length or use continuous states to represent alternative paths. LatentOmni interleaves text with recurrent audio-visual states so dense sensory evidence need not be verbalized \citep{dai2026latentomni}. This motivation is close to ours, but its fixed latent budget and serial recurrence prioritize multimodal grounding over reasoning efficiency and path diversity.
\vspace{-1mm}
\section{Method}
\label{sec:method}
\vspace{-1mm}

\begin{figure}[t]
\centering
\includegraphics[width=\textwidth]{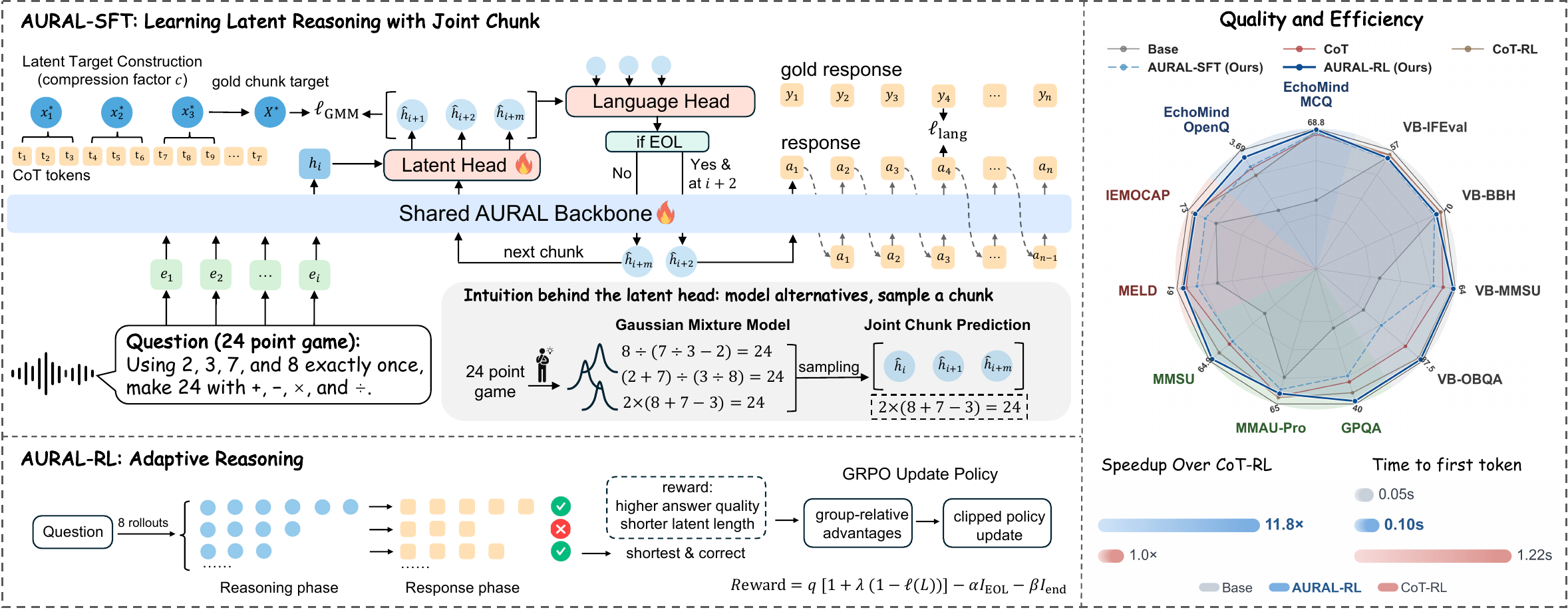}
\caption{Overview of AURAL. \textbf{Top left:} AURAL-SFT pools CoT embeddings into latent targets under compression factor $c$, then predicts a Gaussian mixture over a joint chunk of $m$ future states that reenters the backbone in one pass, while the language head keeps each state readable and emits \textsc{EOL}. \textbf{Bottom left:} AURAL-RL scores 8 rollouts per prompt with a quality-gated conciseness reward, making latent depth adaptive. \textbf{Right:} AURAL-RL is comparable to CoT-RL over eleven metrics while cutting time to first token from $1.22$ to $0.10$\,s ($11.8\times$).}
\label{fig:overview}
\vspace{-2mm}
\end{figure}

\paragraph{Overview.}
Explicit CoT provides direct supervision, but inference must verbalize every intermediate step as a discrete token, making it slow and restricting reasoning to token-expressible paths. AURAL retains CoT only as a training signal. After CoT finetuning on AuralReason-683K (Section~\ref{sec:data_construction}), AURAL-SFT pools consecutive CoT embeddings into continuous targets and uses a Gaussian mixture model (GMM) to represent multiple chunks of future states; at inference, each sampled chunk is fed back in one backbone pass. AURAL-RL then optimizes complete rollouts, adapting reasoning depth to the problem while improving answer quality. Figure~\ref{fig:overview} summarizes both stages.

\subsection{AURAL-SFT: Learning Latent Reasoning with Joint Chunk}
\label{sec:method_latent}
\vspace{-1mm}

\paragraph{From CoT tokens to latent targets.}
The central challenge in supervised latent reasoning is that intermediate latent states have no natural targets. We therefore adopt the simple supervision design of CoLaR \citep{colar2025}, which pools consecutive CoT token embeddings into an aligned latent sequence. This preserves the order of the original CoT and gives every latent position an explicit target, allowing efficient training with teacher forcing. Given a compression factor $c$, let $\mathcal{G}_i=(y_{i,1},\ldots,y_{i,n_i})$ be the $i$th group with embeddings $e_{i,j}\in\mathbb{R}^{d}$, where $n_i=c$ except possibly for the final group. Each group yields one target
\begin{equation}
\label{eq:pool}
u_i=\frac{1}{\sqrt{n_i}}\sum_{j=1}^{n_i}e_{i,j},
\qquad
x_i=\frac{u_i}{s_e},
\end{equation}
where $s_e$ is the embedding standard deviation. The factor $1/\sqrt{n_i}$ preserves scale. This maps $T$ CoT tokens to $\lceil T/c\rceil$ latent targets, while $x_i$ stabilizes density fitting. 

\paragraph{Modeling alternative latents.}
Let $h_i$ denote the final backbone hidden state after processing the question and the latent prefix through position $i$. The intuitive baseline regresses one next state $x_{i+1}$ from $h_i$. Because each training example contains one recorded CoT, point regression treats its pooled state as the unique continuation of that prefix. It can compress the observed trace, but cannot represent uncertainty when the current reasoning state admits several valid next moves. Continuous latent reasoning is useful precisely because it can retain these alternatives before committing to one explicit path \citep{hao2025coconut}. These alternatives may represent different reasoning operations, so their latent targets can occupy separated regions rather than vary around one center. We represent these regions as distinct modes of $p_\theta(x_{i+1}\mid h_i)$, allowing the transition distribution to place probability mass on several alternatives. The GMM lets different components capture different modes, while their weights express the relative plausibility of these continuations. Let $D_{ik}=\operatorname{diag}(\sigma_{ik}^{2})$.
\begin{equation}
\label{eq:gmm}
p_\theta(x_{i+1}\mid h_i)=\sum_{k=1}^{K}\pi_{ik}\,
\mathcal{N}\!\left(x_{i+1}\mid\mu_{ik},D_{ik}\right),
\qquad
\mathcal{L}_{\mathrm{GMM}}(i)=-\frac{1}{d}\log p_\theta(x_{i+1}\mid h_i).
\end{equation}
Here $K$ is the number of Gaussian components and $k$ indexes one component. The loss $\mathcal{L}_{\mathrm{GMM}}$ is the negative log likelihood of the observed pooled target under the predicted mixture, averaged over its $d$ dimensions. Minimizing it jointly learns $\{\pi_{ik},\mu_{ik},\sigma_{ik}^{2}\}_{k=1}^{K}$. Appendix~\ref{app:mode_intervention} intervenes on this component index at a fixed state and traces the effect on the resulting reasoning path.

\paragraph{Joint chunk prediction adds structure and speed.}
The GMM restores multiple possible transitions, but applying it independently at each position still leaves the structure among neighboring latent states implicit. Under teacher forcing, each prediction receives the gold prefix, so independent one-state losses do not directly teach several future states to evolve together. Inference is also serial because each predicted state must pass through the backbone before the next state can be generated. Generating a latent trace of length $L$ therefore requires $L$ sequential forward passes through the backbone. We address both limitations by predicting a chunk of $m$ future states jointly, reducing this cost to $\lceil L/m\rceil$ sequential forward passes. Our latent head uses $m$ learned slot embeddings, one for each future state. A shared MLP maps $h_i$ into the slot feature space. Token mixing exchanges information among slots, while channel mixing refines each slot. Separate linear heads then output $\{\pi_{ik},\mu_{ik},\sigma_{ik}^{2},F_{ik}\}_{k=1}^{K}$. This shared slot design explicitly models relations within the chunk and uses fewer parameters than a monolithic MLP that emits the entire chunk. Formally
\begin{equation}
\label{eq:chunk}
\begin{aligned}
X_i&=(x_{i+1},\ldots,x_{i+m}), \qquad
p_\theta(X_i\mid h_i)=\sum_{k=1}^{K}\pi_{ik}
\mathcal{N}\!\left(\operatorname{vec}(X_i)\mid\mu_{ik},\Sigma_{ik}\right),\\
\Sigma_{ik}&=\operatorname{diag}(\sigma_{ik}^{2})+F_{ik}F_{ik}^{\top},
\qquad F_{ik}\in\mathbb{R}^{md\times r}, \qquad r=16.
\end{aligned}
\end{equation}
The operator $\operatorname{vec}(X_i)\in\mathbb{R}^{md}$ concatenates the $m$ latent states into one vector. The diagonal term captures coordinate uncertainty, while the low-rank term couples positions within the chunk. We use a masked negative log likelihood for partial final chunks.

\paragraph{CSA keeps latent reasoning connected to language.}
The freedom introduced by the GMM chunk distribution creates a complementary risk. Latent states may drift far away from language space, yet the language head must eventually predict a special \textsc{EOL} token that stops latent reasoning and starts response generation. If this connection is learned only at the boundary, both the stopping decision and answer decoding become difficult. We therefore use chunkwise semantic alignment (CSA) as a soft language anchor throughout reasoning
\begin{equation}
\label{eq:csa}
\mathcal{L}_{\mathrm{CSA}}(i)
=-\frac{1}{n_i}\sum_{j=1}^{n_i}\log p_\theta(y_{i,j}\mid h_{i-1}).
\end{equation}
When $h_{i-1}$ predicts group $\mathcal{G}_i$, the language head maps it to a distribution over the vocabulary. CSA assigns equal target mass to all $n_i$ CoT tokens summarized by this group. It therefore keeps each latent state semantically readable by the language head, making the \textsc{EOL} decision and final response easier to decode.

\paragraph{LSS reduces exposure bias in continuous space.}
CSA anchors latent states to language, but teacher forcing still creates a mismatch. Training consumes gold latent states, whereas inference consumes samples from the GMM, so early errors can move later states outside the training distribution. We address this with latent scheduled sampling (LSS), a two-pass adaptation of scheduled sampling \citep{bengio2015scheduled,mihaylova2019scheduled}. The first pass predicts reparameterized chunks from gold inputs. The second pass consumes these samples and applies the same CSA and token losses. Since the samples are not detached, gradients also train the latent head through predicted inputs. Let $\mathcal{L}_{\mathrm{lang}}$ combine CSA with standard cross entropy on \textsc{EOL} and response tokens. Superscripts $\mathrm{gold}$ and $\mathrm{pred}$ identify the latent inputs used by each pass. The complete objective is
\begin{equation}
\label{eq:latent_objective}
\mathcal{L}_{\mathrm{latent}}
=\mathcal{L}_{\mathrm{GMM}}
+\mathcal{L}_{\mathrm{lang}}^{\mathrm{gold}}
+w_t\mathcal{L}_{\mathrm{lang}}^{\mathrm{pred}},
\qquad w_t=\min(1,t/T_{\mathrm{warm}}).
\end{equation}
Here $t$ is the current optimization step and $T_{\mathrm{warm}}=500$ is the number of warmup updates. Thus $w_t$ increases the predicted input loss linearly from $0$ to $1$ during warmup and remains at $1$ afterward.

\subsection{AURAL-RL: Adaptive Reasoning with Reinforcement Learning}
\label{sec:method_rl}
\vspace{-1mm}

\paragraph{From alternative paths to adaptive depth.}
AURAL-SFT imitates one recorded CoT and its length, so an alternative latent path receives no credit even if it produces the correct answer. AURAL-RL instead scores complete rollouts by their final outcomes, allowing the model to discover valid latent paths and allocate different amounts of depth to different problems. The GMM lets us evaluate the same sampled chunk under the updated policy and the policy that generated it. Because chunk log density scales with the number of accepted latent coordinates, we divide each log density by that number before exponentiating their difference. This normalization keeps latent-action updates comparable across chunk sizes and provides the multiplier for our GRPO-style clipped objective \citep{shao2024deepseekmath}; Appendix~\ref{app_latent_policy} gives the equations. For each prompt, we sample 8 rollouts, compute group-relative advantages, and update all parameters without a reference policy.

\paragraph{Reward design.}
For questions with a verifiable answer, we compute quality $q\in[0,1]$ using exact match or the task-specific checker. Open-ended responses are judged by DeepSeek-V4-Flash, while tasks whose correctness depends on the source audio are judged by Gemini-3.5-Flash. Let $I_{\mathrm{EOL}}$ equal $1$ when a rollout never emits \textsc{EOL}, and let $I_{\mathrm{end}}$ equal $1$ when its response lacks the end token. Both are $0$ otherwise. For latent length $L$ and rollout cap $L_{\max}$, we define the normalized length cost and reward as
\begin{equation}
\label{eq:reward}
\ell(L)=\frac{L-1}{L_{\max}-1},
\qquad
R=q\left[1+\lambda\left(1-\ell(L)\right)\right]
-\alpha I_{\mathrm{EOL}}-\beta I_{\mathrm{end}},
\end{equation}
where $L_{\max}=64$ is the maximum latent budget used for each rollout. The value $\ell(L)$ increases from $0$ for the shortest trace to $1$ at this limit. Thus the conciseness bonus is largest for a short correct trace and decreases linearly with length. Multiplication by $q$ prevents an incorrect answer from earning a shortcut reward by stopping early. This quality-gated bonus is what makes AURAL-RL adaptive: additional latent depth is rewarded only when it contributes to a better outcome. We use $\lambda=0.5$, $\alpha=0.1$, and $\beta=0.02$, with the last two coefficients penalizing invalid termination.

\vspace{-1mm}
\section{Data Construction}
\label{sec:data_construction}
\vspace{-1mm}

Latent reasoners learn their initial behavior from explicit CoT, but existing public speech corpora do not provide reasoning that uses the voice rather than only the transcript. We therefore build AuralReason-683K, a bilingual corpus of 683K utterances and about 1,000 hours of speech for emotion recognition, empathetic dialogue, and general reasoning. 
Appendix~\ref{app:data_construction} gives the full pipeline, per-source counts, and teacher prompts.

To teach the model what the voice can change, we begin with two complementary sources. LIME-440K offers mostly synthesized bilingual affective dialogue, while EmotionCoT-35K supplies English human recordings for emotion classification \citep{zhao2026cogaudio,wang2026emotionthinker}. We retain only their audio and write new CoTs and responses. Controlled speech makes affect easy to isolate, but it covers only a narrow slice of real interaction. We therefore build HumanSpeech-1M from about 20K hours of public videos, podcasts, films, and speech corpora. Scale alone is not sufficient. Useful supervision requires utterances whose vocal delivery materially changes the appropriate assistant response. After removing synthetic or unreliable audio, we screen the remaining clips for this criterion, retaining 75K of 1.69M candidates. These three branches teach voice-sensitive interaction, but they do not teach broad problem solving. GeneralSpeech fills that gap with bilingual tasks in mathematics, knowledge, comprehension, and instruction following. We rewrite public text problems as natural spoken questions before synthesizing them with Qwen3-TTS \citep{hu2026qwen3tts}.

Once the audio is selected, the next question is how much reasoning to write. AURAL derives its latent targets from explicit CoT, so a verbose teacher trace is not harmless. Repeated explanations can become extra latent computation. We therefore seek the shortest trace that still supports the response, retaining every necessary inference while removing restatement and reasoning after the conclusion. Prior work likewise finds that shorter valid CoTs can preserve or improve accuracy \citep{hassid2025overthink,wu2025moreisless,xu2025chainofdraft}. For dialogue, the trace must explain how the voice changes the reply rather than merely name an emotion. Gemini-3.5-Flash listens to each clip and writes states with distinct roles. \emph{Cue} records what is audible, \emph{Need} interprets what the speaker needs, optional \emph{Risk} handles ambiguity or safety, and \emph{Plan} determines the reply. A shaky voice with long pauses, for example, motivates reassurance before advice. Emotion traces instead use audible cues to rule out the closest competing emotion before describing the affective state. Across all tasks, the CoT stops before the final answer so that it teaches a reasoning path rather than a shortcut to the target \citep{shen2025codi}. 

GeneralSpeech begins with a CoT and response from Qwen3-Omni-Thinking \citep{xu2025qwen3omni}. Gemini-3.5-Flash then creates two versions because training and analysis require different guarantees. For the controlled comparison in Section~\ref{sec:exp_analysis}, Gemini shortens only the CoT and keeps the answer unchanged, even when wrong. Any performance difference therefore comes from the rewritten CoT rather than answer correction. For the main training set, it verifies the answer and repairs the CoT and response when needed. Only verified outputs enter the corpus. After filtering and rewriting, LIME-440K, EmotionCoT-35K, HumanSpeech-1M, and GeneralSpeech contain 320K, 18K, 75K, and 270K examples. Their 377K Chinese and 306K English utterances form AuralReason-683K.

\vspace{-2mm}
\section{Experiments}
\label{sec:experiment}
\vspace{-1mm}

For latent reasoning to be practically useful, its efficiency gains must not come at the cost of broad task competence or architectural generality. Our experiments examine whether AURAL preserves answer quality across diverse speech and text tasks, transfers across backbone architectures, and reduces the computation required before producing an answer.

\subsection{Experimental Setup}
\label{sec:exp_setup}
\vspace{-1mm}

We assemble an evaluation suite that covers speech-specific capabilities and broader reasoning. EchoMind \citep{zhou2025echomind} evaluates vocal understanding, integrated reasoning, and empathetic response generation. We report multiple-choice accuracy and the mean judge score across its four official dimensions for empathetic responses. IEMOCAP \citep{busso2008iemocap} and MELD \citep{poria2019meld} evaluate speech emotion recognition on their standard test sets. MMSU \citep{wang2026mmsu} covers fine-grained spoken language perception and reasoning. MMAU-Pro \citep{kumar2026mmaupro} measures broad audio intelligence, while GPQA \citep{rein2024gpqa} measures expert-level scientific reasoning. VoiceBench \citep{chen2026voicebench} spans multiple dimensions of voice assistant capability. We select four subsets with objective metrics that assess knowledge, reasoning, and instruction following. These subsets are OpenBookQA, MMSU, BBH, and IFEval.

Qwen2.5-Omni-7B \citep{xu2025qwen25omni} is the primary backbone. We compare direct answering, explicit CoT, existing latent-reasoning baselines, and AURAL, together with RL-trained variants where applicable. The external latent baselines are Coconut \citep{hao2025coconut}, CODI \citep{shen2025codi}, and CoLaR \citep{colar2025}. All RL variants use the same prompts, schedule, and reward, with length counted in latent states or CoT tokens. Kimi-Audio \citep{kimiteam2025kimiaudio} provides a second architecture for testing transfer. Unless stated otherwise, AURAL uses compression factor $c=4$, chunk size $m=10$, $4$ Gaussian components, and covariance rank 16. Detailed training, evaluation, and timing settings are provided in Appendix~\ref{app:experimental_details}.

\begin{table}[t]
    \caption{Main results across the full evaluation suite. $\mathrm{Latent}_{\mathrm{MSE}}$ and $\mathrm{Latent}_{\mathrm{GMM}}$ respectively regress one next state and model alternative continuations, both without chunking. AURAL-SFT adds joint chunk prediction and AURAL-RL adds adaptive RL. CoT uses two epochs, while latent methods match this budget with one CoT epoch and one latent SFT epoch. OpenQ uses the official 1 to 5 scale. Other entries are percentages, and higher is better.}
    \label{tab:main_results}
    \centering
    \definecolor{tblNavy}{HTML}{243B53}
    \definecolor{tblModelBand}{HTML}{E6ECF0}
    \definecolor{tblStripe}{HTML}{F6F8FA}
    \definecolor{tblAuralSFT}{HTML}{E5F3F0}
    \definecolor{tblAuralRL}{HTML}{CBE7DE}
    \definecolor{tblAuralInk}{HTML}{174E45}
    \definecolor{tblRule}{HTML}{52687A}
    \arrayrulecolor{tblRule}
    \setlength{\tabcolsep}{3.3pt}
    \renewcommand{\arraystretch}{1.15}
    \resizebox{\textwidth}{!}{%
    \begin{tabular}{lccccccccccc}
    \toprule
    & \multicolumn{2}{c}{\textcolor{tblNavy}{\textbf{EchoMind}}}
    & \multicolumn{2}{c}{\textcolor{tblNavy}{\textbf{Emotion}}}
    & \multicolumn{3}{c}{\textcolor{tblNavy}{\textbf{General Ability}}}
    & \multicolumn{4}{c}{\textcolor{tblNavy}{\textbf{VoiceBench}}} \\
    \cmidrule(lr){2-3}\cmidrule(lr){4-5}\cmidrule(lr){6-8}\cmidrule(lr){9-12}
    \textcolor{tblNavy}{\textbf{Method}}
    & \textcolor{tblNavy}{\textbf{MCQ}}
    & \textcolor{tblNavy}{\textbf{OpenQ}}
    & \textcolor{tblNavy}{\textbf{IEMOCAP}}
    & \textcolor{tblNavy}{\textbf{MELD}}
    & \textcolor{tblNavy}{\textbf{MMSU}}
    & \textcolor{tblNavy}{\textbf{MMAU-Pro}}
    & \textcolor{tblNavy}{\textbf{GPQA}}
    & \textcolor{tblNavy}{\textbf{OBQA}}
    & \textcolor{tblNavy}{\textbf{MMSU}}
    & \textcolor{tblNavy}{\textbf{BBH}}
    & \textcolor{tblNavy}{\textbf{IFEval}} \\
    \midrule
    \rowcolor{tblModelBand}
    \multicolumn{12}{l}{\textcolor{tblNavy}{\textbf{Qwen2.5-Omni based methods}}} \\
    Base & 58.09 & 3.492 & 66.24 & 52.87 & 61.60 & 55.66 & 23.44 & 79.78 & 51.53 & 66.70 & 54.64 \\
    \rowcolor{tblStripe}
    CoT & 67.94 & 3.631 & \underline{71.39} & 59.12 & 63.72 & \underline{62.33} & 35.35 & 85.27 & 61.97 & \textbf{68.30} & \underline{55.30} \\
    $\mathrm{Latent}_{\mathrm{Coconut}}$ & 60.83 & 3.548 & 67.12 & 54.29 & 62.08 & 57.41 & 25.82 & 80.66 & 53.87 & 64.50 & 53.47 \\
    \rowcolor{tblStripe}
    $\mathrm{Latent}_{\mathrm{CODI}}$ & 62.37 & 3.532 & 67.85 & 54.06 & 62.40 & 57.82 & 28.21 & 81.32 & 55.30 & 64.20 & 53.92 \\
    $\mathrm{Latent}_{\mathrm{CoLaR}}$ & 65.51 & 3.618 & 68.41 & 55.94 & 63.08 & 58.38 & 31.14 & 81.54 & 58.72 & 64.90 & 53.68 \\
    \rowcolor{tblStripe}
    $\mathrm{Latent}_{\mathrm{MSE}}$ & 65.94 & 3.601 & 68.33 & 55.82 & 62.92 & 58.63 & 31.50 & 81.76 & 58.20 & 64.80 & 53.53 \\
    $\mathrm{Latent}_{\mathrm{GMM}}$ & 67.58 & 3.624 & 68.65 & 56.44 & 63.28 & 59.47 & 33.15 & 81.98 & 59.63 & 65.10 & 53.53 \\
    \rowcolor{tblAuralSFT}
    \textcolor{tblAuralInk}{\textbf{AURAL-SFT}} & 68.20 & 3.637 & 68.82 & 56.97 & 63.54 & 59.91 & 33.88 & 82.20 & 60.38 & 65.00 & 53.81 \\
    \addlinespace[1pt]
    \rowcolor{tblStripe}
    CoT-RL & \underline{68.42} & \underline{3.658} & \textbf{71.80} & \textbf{60.04} & 64.34 & \textbf{62.78} & 37.55 & \underline{87.03} & \underline{63.37} & \underline{68.10} & \textbf{55.52} \\
    $\mathrm{Latent}_{\mathrm{CoLaR}}$-RL & 66.89 & 3.641 & 70.43 & 58.16 & 63.56 & 60.49 & 37.18 & 85.71 & 62.30 & 65.70 & 53.87 \\
    \rowcolor{tblStripe}
    $\mathrm{Latent}_{\mathrm{MSE}}$-RL & 65.72 & 3.623 & 69.86 & 56.40 & 63.42 & 59.70 & 34.62 & 82.20 & 60.41 & 65.40 & 53.17 \\
    $\mathrm{Latent}_{\mathrm{GMM}}$-RL & 67.92 & 3.651 & 70.83 & 59.04 & \underline{64.40} & 61.07 & \underline{38.46} & 86.37 & 62.98 & 65.50 & 53.71 \\
    \rowcolor{tblAuralRL}
    \textcolor{tblAuralInk}{\textbf{AURAL-RL}} & \textbf{68.63} & \textbf{3.668} & 71.31 & \underline{59.69} & \textbf{64.76} & 61.37 & \textbf{39.38} & \textbf{87.25} & \textbf{63.63} & 66.00 & 54.29 \\
    \midrule
    \rowcolor{tblModelBand}
    \multicolumn{12}{l}{\textcolor{tblNavy}{\textbf{Kimi-Audio based methods}}} \\
    Base & 62.76 & 2.614 & 60.19 & 48.54 & 56.54 & 52.35 & 29.30 & 83.30 & 61.71 & 66.20 & 63.07 \\
    \rowcolor{tblStripe}
    CoT & 69.41 & 3.093 & 64.14 & 53.72 & 58.28 & \textbf{57.80} & 35.16 & \underline{87.25} & \textbf{66.53} & 67.10 & 64.32 \\
    CoT-RL & \textbf{70.12} & \underline{3.162} & \textbf{65.27} & \underline{55.40} & \textbf{59.14} & \underline{57.44} & \underline{37.36} & 86.81 & 66.14 & \textbf{67.90} & \textbf{65.04} \\
    \rowcolor{tblAuralSFT}
    \textcolor{tblAuralInk}{\textbf{AURAL-SFT}} & 67.31 & 2.831 & 62.77 & 52.18 & 57.82 & 55.41 & 35.35 & 85.71 & 65.29 & 66.70 & 63.41 \\
    \rowcolor{tblAuralRL}
    \textcolor{tblAuralInk}{\textbf{AURAL-RL}} & \underline{69.53} & \textbf{3.184} & \underline{64.87} & \textbf{56.09} & \underline{58.62} & 56.82 & \textbf{38.64} & \textbf{88.57} & \underline{66.36} & \underline{67.50} & \underline{64.80} \\
    \bottomrule
    \end{tabular}%
    }
    \arrayrulecolor{black}
    \end{table}

\subsection{Main Results}
\label{sec:exp_main}
\vspace{-1mm}

Table~\ref{tab:main_results} compares AURAL-SFT and AURAL-RL with direct answering, explicit CoT, and latent baselines across both backbones. Because CoT provides the targets for latent states, AURAL-SFT inherits its gains rather than surpassing it: on Qwen2.5-Omni it outperforms Base on most measures while remaining below CoT on the majority. AURAL-RL then improves all 11 scores. Coconut and CODI perform worse than the other latent methods on most benchmarks. They lack direct supervision for intermediate states and use a fixed latent length, which may make latent trajectories harder to learn and prevent depth from adapting to question difficulty. AURAL-RL leads both EchoMind metrics and both MMSU evaluations, which rely heavily on spoken and paralinguistic information. The margins are small but consistent with latent states retaining cues that are hard to verbalize.

A closer comparison of the latent variants shows the empirical effects of each AURAL design choice. Compared with $\mathrm{Latent}_{\mathrm{MSE}}$, $\mathrm{Latent}_{\mathrm{GMM}}$ improves ten metrics and ties one, suggesting that modeling multiple plausible next-state continuations is more effective than predicting a single state. AURAL-SFT then improves ten of eleven metrics over $\mathrm{Latent}_{\mathrm{GMM}}$, consistent with an additional benefit from jointly modeling relations among adjacent latent states. This flexibility becomes especially important in AURAL-RL. Its SFT-to-RL gain exceeds CoT's on most metrics under our setup. These larger gains likely arise because outcome rewards allow the distributional latent policy to explore multiple continuous trajectories and reinforce those that yield stronger final answers, rather than remaining tied to the single CoT path used for supervision. Separately, the MSE model gains less from RL than the CoLaR and GMM variants, consistent with distributional policies offering broader exploration than point prediction.

The results on Kimi-Audio align with the overall pattern on Qwen2.5-Omni. AURAL-SFT again lies between Base and CoT on most measures, and AURAL-RL improves all eleven of its scores. AURAL-RL remains competitive with CoT-RL, while its larger benefit from RL recurs across most evaluation metrics. Together, these trends show that AURAL's combination of CoT supervision, distributional latent modeling, joint chunk prediction, and adaptive RL transfers across backbones.

\subsection{Compression Factor and Chunk Size}
\label{sec:exp_grid}
\vspace{-1mm}

\begin{figure}[t]
    \centering
    \includegraphics[width=\textwidth]{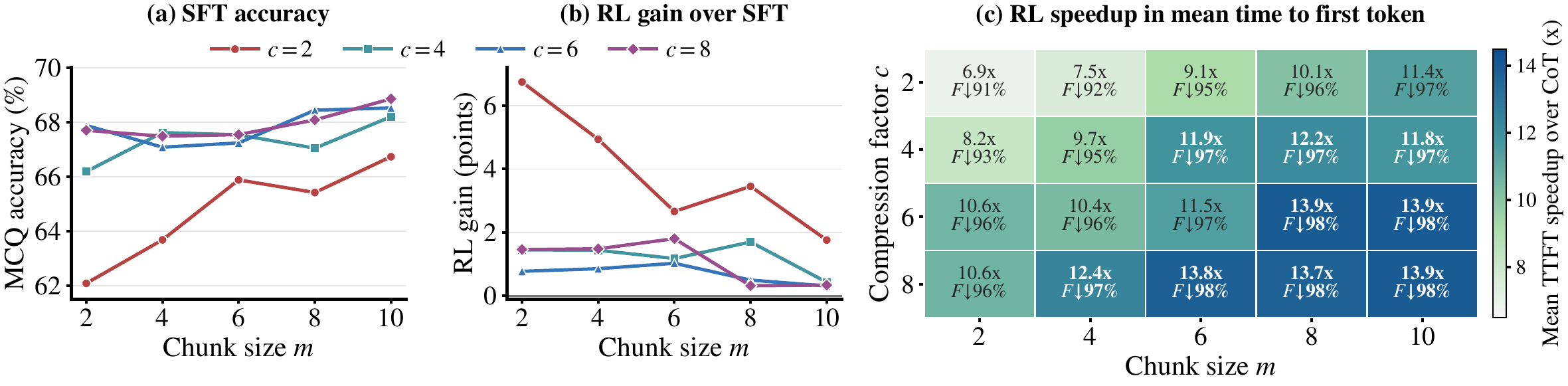}
    \caption{Effects of compression factor $c$, chunk size $m$, and RL on EchoMind MCQ and response latency. Panel a reports AURAL-SFT accuracy. Panel b reports the gain from AURAL-RL over the matching SFT checkpoint. Panel c reports measured AURAL-RL speedup in time to first token over CoT-RL, as a ratio of means under the protocol of Appendix~\ref{app:response_latency}. Color and the value marked with $\times$ encode this speedup, and $F\downarrow$ gives the reduction in serial reasoning forward passes, where each CoT reasoning token and each latent chunk require one pass.}
    \label{fig:rate_chunk_grid}
    \vspace{-2mm}
    \end{figure}

Panel a of Figure~\ref{fig:rate_chunk_grid} examines how the semantic granularity set by $c$ and the joint prediction span set by $m$ affect AURAL-SFT. It shows that $c=2$ performs clearly worse, whereas $c=4$, $6$, $8$ are close. With $c=2$, each target pools only a narrow local fragment and may reflect local wording more than a coherent reasoning unit. Moderate compression aggregates a broader semantic span and gives the latent head more stable targets. The plateau beyond $c=4$ suggests that this abstraction benefit saturates once each state contains sufficient reasoning context. Along the chunk dimension, average accuracy rises almost monotonically with $m$, despite small fluctuations in individual curves. A larger $m$ lets the model learn a longer segment of the latent trajectory jointly, which likely drives the overall upward trend. Panel b further supports this interpretation. AURAL-RL improves every combination, with the largest gains at $c=2$ and similar smaller gains at the other compression factors. Unlike SFT, RL is guided by final answers rather than local CoT targets, so it can compensate most when the supervised states at $c=2$ provide insufficient semantic abstraction. Once $c\geq4$ provides coherent targets, the remaining room for improvement becomes smaller. Appendix~\ref{app:performance_grid} shows the same grid for the other ten evaluation metrics, which follow both patterns.

Although accuracy largely saturates once $c\geq4$, compression and chunking still shape efficiency. Explicit CoT generates one reasoning token in each serial forward pass, whereas AURAL-RL generates an entire latent chunk in one pass. Let $F_{\downarrow}$ denote the resulting reduction in the number $N$ of sequential forward passes
\begin{equation}
\label{eq:reasoning_speedup}
F_{\downarrow}=1-\frac{N_{\mathrm{AURAL\text{-}RL\,chunk}}}{N_{\mathrm{CoT\text{-}RL\,token}}}.
\end{equation}
Panel c reports both $F_{\downarrow}$ and the speedup in mean time to first token over CoT-RL. Mean time to first token covers all model computation before the first visible answer token, including processing the input and performing reasoning; AURAL primarily accelerates reasoning. At $c=4,m=10$, it reduces the mean number of serial reasoning passes from 64.25 token passes to 1.79 chunk passes ($F_{\downarrow}=97.21\%$) and achieves an $11.8\times$ speedup. Although $c=8,m=10$ reduces latent reasoning to one pass, it improves mean time to first token by only $0.016$\,s over the default, showing that most remaining latency lies outside reasoning. Since $c=4$ is the first point on the accuracy plateau, we use $c=4,m=10$ as the default. Appendix~\ref{app:response_latency} provides a full latency breakdown.

\vspace{-1mm}

\subsection{Ablation Study}
\label{sec:exp_ablation}
\vspace{-2mm}

\begin{wraptable}[12]{R}{0.52\textwidth}
\vspace{-\intextsep}
\setlength{\abovecaptionskip}{4pt}
\caption{Ablations of AURAL-RL.}
\label{tab:ablation}
\centering
\definecolor{ablNavy}{HTML}{243B53}
\definecolor{ablStripe}{HTML}{F6F8FA}
\definecolor{ablAural}{HTML}{CBE7DE}
\definecolor{ablAuralInk}{HTML}{174E45}
\definecolor{ablRule}{HTML}{52687A}
\arrayrulecolor{ablRule}
\setlength{\tabcolsep}{3.8pt}
\renewcommand{\arraystretch}{1.12}
\resizebox{\linewidth}{!}{%
\begin{tabular}{lcccc}
\toprule
\textcolor{ablNavy}{\textbf{Variant}}
& \textcolor{ablNavy}{\shortstack{\textbf{EchoMind}\\\textbf{MCQ}}}
& \textcolor{ablNavy}{\shortstack{\textbf{MMSU}\\\textbf{Acc.}}}
& \textcolor{ablNavy}{\shortstack{\textbf{Latent}\\\textbf{states}}}
& \textcolor{ablNavy}{\shortstack{\textbf{EOL hit}\\\textbf{rate}}} \\
\midrule
\rowcolor{ablAural}
\textcolor{ablAuralInk}{\textbf{AURAL-RL}} & 68.63 & 64.76 & 12.36 & 100.00 \\
w/o CSA & \textsc{64.74} & \textsc{63.16} & \textsc{12.58} & \textsc{96.62} \\
\rowcolor{ablStripe}
w/o LSS & \textsc{66.82} & \textsc{64.30} & \textsc{11.92} & \textsc{98.80} \\
w/o conciseness bonus & \textsc{68.81} & \textsc{63.96} & \textsc{14.77} & \textsc{100.00} \\
\addlinespace[1pt]
\rowcolor{ablStripe}
$K=1$ & \textsc{65.99} & \textsc{63.68} & \textsc{13.15} & \textsc{100.00} \\
$K=2$ & \textsc{67.02} & \textsc{63.84} & \textsc{11.94} & \textsc{100.00} \\
\rowcolor{ablStripe}
$K=8$ & \textsc{68.55} & \textsc{64.88} & \textsc{12.40} & \textsc{100.00} \\
$K=16$ & \textsc{68.17} & \textsc{64.26} & \textsc{12.71} & \textsc{99.82} \\
\rowcolor{ablStripe}
$r=0$ & \textsc{66.10} & \textsc{63.58} & \textsc{12.45} & \textsc{99.93} \\
\bottomrule
\end{tabular}%
}
\arrayrulecolor{black}
\end{wraptable}

Table~\ref{tab:ablation} first asks what keeps AURAL's latent rollouts accurate and ensures a reliable transition to answer generation. Removing CSA causes the largest quality loss, reducing EchoMind MCQ by 3.89 points and lowering the \textsc{EOL} hit rate to 96.62\%, while latent depth barely changes. The failure therefore comes not from insufficient reasoning but from intermediate states drifting away from language space, which weakens both answer decoding and the decision to end latent reasoning. Removing LSS causes a smaller but consistent decline in accuracy and termination, which supports its role in preparing the model for states sampled from its own policy.

The quality-gated conciseness bonus $\lambda q[1-\ell(L)]$ in Equation~\ref{eq:reward} rewards shorter correct traces. Removing it raises EchoMind MCQ by 0.18 points but lowers MMSU by 0.80, while the \textsc{EOL} hit rate stays at 100.00\% and latent depth increases by 19.5\%, so the bonus suppresses redundant latent states without a consistent loss in accuracy.

Mixture size $K$ asks how many continuation modes are needed. A single Gaussian loses 2.64 points on EchoMind MCQ and $K=2$ recovers only part of that gap, so one or two modes do not capture enough variation, while raising $K$ to 8 or 16 brings no consistent gain and slightly lengthens the trajectory. Four components therefore suffice. Setting $r=0$ leaves $\Sigma_{ik}$ in Equation~\ref{eq:chunk} diagonal and costs 2.53 points, so coupling positions within a chunk matters beyond per-coordinate uncertainty.

\vspace{-1mm}
\subsection{Analysis}
\label{sec:exp_analysis}
\vspace{-1mm}

\paragraph{Shorter CoTs provide cleaner supervision.}
Section~\ref{sec:data_construction} creates paired GeneralSpeech examples with the same question and answer but either the original or shortened CoT. We fine-tune one Qwen2.5-Omni model on each set with the same setup and evaluate both on GPQA. The answers remain unchanged even when wrong, making the CoT the only difference in supervision. Shortening reduces the mean trace length from 317.5 to 65.3 tokens and raises GPQA accuracy from 29.85\% to 35.35\%. This result supports concise CoTs as a cleaner training signal.

\vspace{-2mm}
\paragraph{Do GMM components represent different valid paths?}
We copy the same intermediate reasoning state four times. The copies share the question and latent prefix, but a different GMM component produces the next latent chunk in each one. AURAL then continues from each chunk to a final answer. This isolates the effect of choosing a component. We sample one component four times to test whether randomness alone creates different paths. We also replace the components with one Gaussian covering the same overall region, asking whether one wide distribution works without separate components. If the original GMM still yields more varied correct paths, its components encode useful alternatives.

\begin{table}[t]
\caption{Testing whether GMM components provide useful alternatives. Each branching row generates four paths. Two valid paths counts prompts with two meaningfully different correct paths. Individual accuracy is the average accuracy of the four paths before voting. GPQA, OBQA, and BBH report majority-vote accuracy. The first row is the standard one-path reference.}
\label{tab:mixture_branching}
\centering
\definecolor{mixNavy}{HTML}{243B53}
\definecolor{mixStripe}{HTML}{F6F8FA}
\definecolor{mixAural}{HTML}{CBE7DE}
\definecolor{mixAuralInk}{HTML}{174E45}
\definecolor{mixRule}{HTML}{52687A}
\arrayrulecolor{mixRule}
\setlength{\tabcolsep}{5pt}
\renewcommand{\arraystretch}{1.12}
\small
\begin{tabularx}{\textwidth}{>{\raggedright\arraybackslash}Xccccc}
\toprule
\textcolor{mixNavy}{\textbf{Generation scheme}}
& \textcolor{mixNavy}{\textbf{Two valid paths$\uparrow$}}
& \textcolor{mixNavy}{\textbf{Individual acc.}}
& \textcolor{mixNavy}{\textbf{GPQA}}
& \textcolor{mixNavy}{\textbf{OBQA}}
& \textcolor{mixNavy}{\textbf{BBH}} \\
\midrule
Standard AURAL-RL, one path & -- & 63.57\% & 39.38\% & 87.25\% & 66.00\% \\
\midrule
\rowcolor{mixAural}
\textcolor{mixAuralInk}{\textbf{One path per GMM component}} & \textbf{38.88\%} & 62.19\% & \textbf{42.49\%} & \textbf{89.23\%} & \textbf{68.30\%} \\
Four samples from one component & 12.84\% & 62.81\% & 40.29\% & 87.69\% & 66.50\% \\
\rowcolor{mixStripe}
Four components merged & 16.09\% & 62.48\% & 40.84\% & 88.13\% & 67.00\% \\
\bottomrule
\end{tabularx}
\arrayrulecolor{black}
\end{table}

Different paths are useful only if their difference exceeds random variation and both answer correctly. Table~\ref{tab:mixture_branching} reports how often a question meets both conditions. Nearly four in ten questions meet this criterion when the four components are used. This is about three times as frequent as sampling one component and more than twice as frequent as merging the components. Individual accuracy changes little, so the difference cannot be explained by extra noise. Voting provides a second test. It selects the most frequent final answer among the four paths and is used only as a diagnostic. If the components are redundant, their vote should be no better than voting over four samples from one component. Instead, it performs better on all three benchmarks, showing that the component paths make complementary errors. Together with the $K=1$ ablation in Table~\ref{tab:ablation}, these results suggest that the GMM keeps plausible next steps separate and gives RL more than one route to a correct answer. Appendix~\ref{app:mode_intervention} gives the exact protocol.

\vspace{-2mm}
\paragraph{Harder questions elicit deeper latent reasoning.}
AURAL-RL's adaptive stopping rule is useful only if it reserves depth for questions that need it. To separate task difficulty from speech properties, we analyze GPQA and VoiceBench BBH, whose solutions rely little on paralinguistic cues. DeepSeek-V4-Flash rates the difficulty of each question on a scale from 1 to 5 without seeing the benchmark identity, latent trajectory, model answer, or reference. Mean latent depth rises from 13.27 to 15.07 across GPQA difficulty levels 3 to 5 and from 10.68 to 13.47 across BBH levels 1 to 3. Levels with too few examples are omitted. The same progression on both benchmarks indicates that AURAL-RL allocates more depth to harder questions rather than using a fixed budget.

\vspace{-2mm}
\section{Conclusion}
\label{sec:conclusion}
\vspace{-2mm}

We introduced AURAL, a framework for SLMs that replaces token-by-token CoT at inference with continuous latent chunks, models multiple plausible continuations with a GMM, and uses RL to adapt reasoning depth. We also constructed AuralReason-683K, a bilingual corpus of 683K speech utterances, and showed across two SLM backbones that AURAL-RL remains competitive with CoT while achieving an $11.8\times$ speedup in mean time to first token. Overall, AURAL brings deep reasoning closer to the immediacy required for spoken interaction.
\section*{AI Use Statement}

During the preparation of this manuscript, a Large Language Model (LLM) was utilized as a writing aid to improve the overall linguistic quality and clarity. This assistance was confined to copy-editing tasks, such as correcting grammatical and spelling errors, rephrasing sentences for enhanced flow and readability, and ensuring conciseness. All scientific contributions, including the research ideas, experimental design, analysis, and conclusions presented herein, are entirely the original work of the human authors.

\section*{Ethics Statement}

This work constructs a speech reasoning corpus from public videos, podcasts, films, and existing speech datasets, and trains models to reason about paralinguistic cues. Inferring affect from voice can shade into profiling, so CoT supervision is limited to acoustic impressions that bear on the response. Perceived age, gender, or accent may appear only in this form, for example when a child's voice calls for simpler wording. Traces describe how the voice sounds rather than who the speaker is, and inferences about stable personality, illness, and social status are excluded. Crisis, medical, and self harm content receives dedicated safety handling. The audio carries the licenses of its sources, several of which permit research use only, so redistribution follows the terms of each source and uses permitted subsets with source identifiers where waveform redistribution is restricted. No new human subjects were recruited, and because the recorded speakers did not consent to this specific use, we treat that as a reason to limit release rather than broaden it. Reasoning traces are generated by teacher models and may carry their biases, including biased impressions of age, gender, or accent.

\section*{Reproducibility Statement}

Appendix~\ref{app_latent_policy} specifies the latent head and the training objectives. Appendix~\ref{app:experimental_details} gives the backbones, training schedules, and evaluation and decoding settings; every setting uses seed 42 and a single training run. Appendix~\ref{app:response_latency} gives the latency measurement protocol, and Appendix~\ref{app:data_construction} documents the corpus and the teacher prompts. We will release the training and evaluation code on publication, together with the data manifests and processing scripts needed to reconstruct the corpus from its sources, subject to the license constraints noted in the Ethics Statement.

\bibliography{iclr2027_conference}
\bibliographystyle{iclr2027_conference}

\clearpage
\appendix
\raggedbottom
\renewcommand{\floatpagefraction}{0.8}
\newcolumntype{Y}{>{\raggedright\arraybackslash}X}
\setlist[itemize]{leftmargin=*,nosep}
\setlist[enumerate]{leftmargin=*,nosep}

\section{AURAL Head and Training Details}
\label{app_latent_policy}

We specify the AURAL-specific latent head and training choices here. Let $d$ be the backbone width, $d_h$ the head width, $m$ the number of slots per chunk, $K$ the number of mixture components, and $r$ the covariance rank.

\subsection{Joint Chunk Parameterization}

At anchor $i$, a shared trunk maps the backbone state $h_i\in\mathbb{R}^{d}$ to one context feature. Learned relative-slot embeddings then distinguish the $m$ future positions:
\begin{equation}
\begin{aligned}
z_i&=\operatorname{LN}_{\mathrm{tr}}
\!\left(W_2\phi(W_1h_i+b_1)+b_2\right),\\
S_i^{(0)}&=\mathbf{1}_m z_i^\top+E,
\qquad E\in\mathbb{R}^{m\times d_h}.
\end{aligned}
\end{equation}
Here $\phi$ is GELU. A token mixer communicates across rows, followed by a channel mixer shared across slots:
\begin{equation}
\begin{aligned}
S_i^{(1)}
&=S_i^{(0)}+W_{\mathrm{tok}}
\operatorname{LN}_{\mathrm{tok}}(S_i^{(0)}),\\
S_i^{(2)}
&=S_i^{(1)}
+\phi\!\left(
\operatorname{LN}_{\mathrm{ch}}(S_i^{(1)})W_{\mathrm{ch},1}
+\mathbf{1}_m b_{\mathrm{ch},1}^{\top}
\right)W_{\mathrm{ch},2}
+\mathbf{1}_m b_{\mathrm{ch},2}^{\top}.
\end{aligned}
\end{equation}
The token-mixing output and final channel projection are zero initialized, so this residual mixer initially preserves the slot features. Write $q_{i,s}$ for row $s$ of $S_i^{(2)}$.

The chunk-level logits are $\alpha_i=W_\pi z_i+b_\pi$, and the weights are
$\pi_i=\operatorname{softmax}(\alpha_i)$, so one component is selected for the entire chunk. At each slot, let $W_{\bullet,k}$ and $b_{\bullet,k}$ denote the $k$th output block of the corresponding shared map. The component parameters are
\begin{equation}
\begin{aligned}
\mu_{iks} &= W_{\mu,k}q_{i,s}+b_{\mu,k},\\
\eta_{iks} &= W_{\eta,k}q_{i,s}+b_{\eta,k},\\
a_{iks} &= W_{a,k}q_{i,s}+b_{a,k},\\
\sigma_{iks} &=
\exp\!\left(\operatorname{clip}
(\eta_{iks},\eta_{\min},\eta_{\max})\right).
\end{aligned}
\end{equation}
Thus $\mu_{iks},\sigma_{iks}\in\mathbb{R}^{d}$ are the slotwise mean and coordinate scale, while $a_{iks}\in\mathbb{R}^{r}$ is its loading vector.

The model learns global directions $B\in\mathbb{R}^{d\times r}$ shared by all anchors, components, and slots. Let the slot-$s$ block row of $F_{ik}$ be
\begin{equation}
F_{ik}^{(s)}=B\operatorname{Diag}(a_{iks}),
\end{equation}
and let $D_{ik}$ be diagonal with slot block $\operatorname{Diag}(\sigma_{iks}^{2})$. Each component has the low-rank-plus-diagonal covariance
\begin{equation}
\begin{aligned}
\Sigma_{ik}&=D_{ik}+F_{ik}F_{ik}^{\top},\\
[\Sigma_{ik}]_{s,t}
&=\mathbb{I}[s=t]\operatorname{Diag}(\sigma_{iks}^{2})
+B\operatorname{Diag}(a_{iks}\odot a_{ikt})B^\top .
\end{aligned}
\end{equation}
The off-diagonal slot blocks couple stochastic deviations across future positions; token mixing instead couples the deterministic parameters used to define those deviations. We use $\mu_{ik}\in\mathbb{R}^{md}$ for the component means concatenated in slot order.

\subsection{Likelihood, Sampling, and Feedback}

For a partial final chunk, let $\mathcal{S}_i\subseteq\{1,\ldots,m\}$ contain the valid slots and let $P_i$ restrict a chunk to those coordinates. With $d_i^v=d|\mathcal{S}_i|$, define
\begin{equation}
\begin{aligned}
Y_i^\star
&=P_i\operatorname{vec}(X_i^\star),\\
\mu_{ik}^v
&=P_i\mu_{ik},\\
D_{ik}^v
&=P_iD_{ik}P_i^\top,\\
F_{ik}^v
&=P_iF_{ik},\\
\Sigma_{ik}^v
&=D_{ik}^v+F_{ik}^v(F_{ik}^v)^\top .
\end{aligned}
\end{equation}
This is the exact Gaussian marginal on the valid slots, with the original mixture weights; it is not zero padding and assigns no density cost to nonexistent targets. The supervised mixture density and the chunk form of $\mathcal{L}_{\mathrm{GMM}}$ in Equation~\ref{eq:latent_objective} are
\begin{equation}
\begin{aligned}
p_\theta^v(Y_i^\star\mid h_i)
&=\sum_{k=1}^{K}\pi_{ik}
\mathcal{N}(Y_i^\star\mid\mu_{ik}^v,\Sigma_{ik}^v),\\
\mathcal{L}_{\mathrm{GMM}}
&=-\frac{1}{|\mathcal{A}|}
\sum_{i\in\mathcal{A}}
\frac{\log p_\theta^v(Y_i^\star\mid h_i)}{d_i^v},
\end{aligned}
\end{equation}
where $\mathcal{A}$ is the set of supervised anchors. Each target $Y_i^\star$ is detached for this loss, and normalization by $d_i^v$ makes full and partial chunks comparable per valid coordinate.

The implementation evaluates each component in FP32 through Woodbury reduction and a rank-$r$ Cholesky factorization, without materializing the full covariance.

For sampling and policy scoring, temperature is part of the distribution rather than only the draw. For any retained slot set $\mathcal{S}$ and $\tau>0$, define the tempered density
\begin{equation}
\begin{aligned}
\pi_{ik}^{(\tau)}
&=\left[\operatorname{softmax}(\alpha_i/\tau)\right]_k,\\
p_{\theta,\tau}^{\mathcal{S}}(Y\mid h_i)
&=\sum_{k=1}^{K}\pi_{ik}^{(\tau)}
\mathcal{N}\!\left(
Y\mid\mu_{ik}^{\mathcal{S}},
\tau^2\Sigma_{ik}^{\mathcal{S}}
\right).
\end{aligned}
\end{equation}
The supervised density above is the $\tau=1$ case. With independent coordinate noise $\epsilon_{i,s}\sim\mathcal{N}(0,I_d)$ and factor noise $g_i\sim\mathcal{N}(0,I_r)$ shared by every slot, a stochastic joint chunk is sampled as
\begin{equation}
\begin{aligned}
\kappa_i
&\sim\operatorname{Categorical}(\pi_i^{(\tau)}),\\
\widehat{x}_{i,s}&
=\mu_{i\kappa_i s}
+\tau\!\left[
\sigma_{i\kappa_i s}\odot\epsilon_{i,s}
+B(a_{i\kappa_i s}\odot g_i)
\right].
\end{aligned}
\end{equation}
The shared $g_i$ realizes the cross-slot covariance while retaining slot-specific loadings.

Samples live in normalized target space and are rescaled before feedback, $\widetilde{x}_{i,s}=s_e\widehat{x}_{i,s}$. All slots enter the backbone in one cached forward call; causal masking preserves their order and lets each returned state attend to earlier accepted slots. The language head evaluates \textsc{EOL} at every returned state. If the first accepted stop is at $s^\star$, only the prefix through $s^\star$ remains and all later slot states and key--value cache entries are truncated. Otherwise the final state anchors the next chunk prediction.

\textsc{EOL} is a special vocabulary token. At every returned latent state, the stop probability is the language-head probability of \textsc{EOL}, and the stop action is a Bernoulli draw at temperature $\tau_{\mathrm{EOL}}=1.0$ in both RL rollouts and evaluation. After the first stop, the model consumes one fixed closing token, which we call the close anchor, and answer decoding starts from the resulting state.

\subsection{AURAL-SFT and AURAL-RL Implementation}

\paragraph{Latent scheduled sampling.}
LSS follows the two-pass procedure in Section~\ref{sec:method_latent}. A teacher-forced pass first computes hidden states at every gold latent position. The joint head is queried only at the stride-$m$ anchors used by decoding, so its sampled chunks tile the latent inputs; the last chunk uses only its valid slots. A second backbone pass replaces the gold latent inputs by these samples and applies the same CSA, \textsc{EOL}, and response objectives. The latent head is still queried at gold anchors, so LSS trains the backbone and language head on sampled inputs, while drift across successive chunks is left to AURAL-RL rollouts.

The sampled component index $\kappa_i$ is hard and treated with stop-gradient. The reparameterized continuous sample is \emph{not} detached, so the second-pass language loss updates the selected means, scales, loadings, global directions $B$, and the backbone through feedback; the supervised mixture NLL supplies the direct signal for the categorical mixture and all components.

\paragraph{Reinforcement learning.}
Each rollout contains three action classes:
\begin{equation}
\begin{aligned}
\mathcal{T}^{\mathrm{lat}}
&=\{\text{accepted latent chunks}\},\\
\mathcal{T}^{\mathrm{stop}}
&=\{\text{binary stop events}\},\\
\mathcal{T}^{\mathrm{ans}}
&=\{\text{answer tokens}\}.
\end{aligned}
\end{equation}
The sampled component index is internal to the sampler, not a separate action. If stopping retains slots $\mathcal{S}_i^{\mathrm{acc}}$, the latent action is $\widehat{Y}_i=P_i^{\mathrm{acc}}\operatorname{vec}(\widehat{X}_i)$ and is scored by its marginal mixture density:
\begin{equation}
\ell_{\theta,\tau,i}^{\mathrm{lat}}
=\frac{1}{d_i^v}
\log p_{\theta,\tau}^{\mathcal{S}_i^{\mathrm{acc}}}
(\widehat{Y}_i\mid h_i),
\qquad d_i^v=d|\mathcal{S}_i^{\mathrm{acc}}|.
\end{equation}
This is a dimension-normalized surrogate log score, chosen empirically to make scores comparable across accepted chunk sizes. Consequently,
\begin{equation}
\widetilde{\rho}_i^{\mathrm{lat}}
=\exp\!\left(
\ell_{\theta,\tau,i}^{\mathrm{lat}}
-\ell_{\theta_{\mathrm{old}},\tau,i}^{\mathrm{lat}}
\right)
\end{equation}
is not a strict importance ratio: dividing each chunk log density by $d_i^v$ changes the exponent. We use it only as a length-comparable surrogate policy multiplier.

When $\tau\neq1$, both the behavior score and current-policy score are evaluated under the actual tempered distributions $p_{\theta_{\mathrm{old}},\tau}$ and $p_{\theta,\tau}$ defined above; neither score is recomputed from untempered weights or covariance. Stop actions use the Bernoulli \textsc{EOL} log probability and answer actions use the categorical token log probability. Their exponentiated current-minus-behavior log-probability differences are standard policy ratios, with stop events included only through the first halt.

Rollout reward $R_g$ is taken directly from Equation~\ref{eq:reward}. For the $G=8$ rollouts of a prompt, rollout $g$ receives $\widehat{A}_g=(R_g-\operatorname{mean}_{g'}R_{g'})/(\operatorname{std}_{g'}R_{g'}+10^{-6})$, shared by all of its actions. For each class $c\in\{\mathrm{lat},\mathrm{stop},\mathrm{ans}\}$, let $\mathcal{T}_g^{c}$ contain the class-$c$ actions of rollout $g$. The class loss averages first within each rollout and then over rollouts:
\begin{equation}
\mathcal{L}_{c}^{\mathrm{clip}}
=-\frac{1}{G}\sum_{g=1}^{G}\frac{1}{|\mathcal{T}_g^{c}|}
\sum_{t\in\mathcal{T}_g^{c}}
\min\!\left(\rho_{g,t}\widehat{A}_g,\,
\operatorname{clip}(\rho_{g,t},1-\epsilon_{\mathrm{clip}},1+\epsilon_{\mathrm{clip}})\widehat{A}_g\right),
\end{equation}
where $\rho_{g,t}$ is $\widetilde{\rho}^{\mathrm{lat}}$ for latent chunks and the strict policy ratio otherwise, and $\epsilon_{\mathrm{clip}}=0.2$ is the clipping coefficient in Table~\ref{tab:rl_details}. With symbolic class weights, the total policy loss for a prompt is
\begin{equation}
\mathcal{L}_{\mathrm{AURAL\text{-}RL}}
=\omega_{\mathrm{lat}}\mathcal{L}_{\mathrm{lat}}^{\mathrm{clip}}
+\omega_{\mathrm{stop}}\mathcal{L}_{\mathrm{stop}}^{\mathrm{clip}}
+\omega_{\mathrm{ans}}\mathcal{L}_{\mathrm{ans}}^{\mathrm{clip}},
\end{equation}
and the batch loss averages it over prompts. We take one gradient step per rollout batch, so $\theta=\theta_{\mathrm{old}}$ when the loss is evaluated. Every ratio then equals one up to numerical differences, clipping is inactive for all three classes, and the update reduces to a group-relative policy gradient. For a latent action this gradient is $\nabla_\theta\ell_{\theta,\tau,i}^{\mathrm{lat}}=\nabla_\theta\log p_{\theta,\tau}^{\mathcal{S}_i^{\mathrm{acc}}}(\widehat{Y}_i\mid h_i)/d_i^v$, so the normalization sets the scale of the latent term relative to the stop and answer terms. In particular, at the training temperature $\tau=1$ the mixture logits receive the direct gradient $(\gamma_{ik}-\pi_{ik})/d_i^v$, where $\gamma_{ik}$ is the component posterior, so the direct RL signal on the mixture weights is small. Training uses no learned critic, reference policy, or KL regularizer.

\section{Experimental Details}
\label{app:experimental_details}

\begingroup
\setlength{\tabcolsep}{6pt}
\renewcommand{\arraystretch}{1.12}
\setlength{\belowcaptionskip}{4pt}
\makeatletter
\patchcmd{\@makecaption}{#1: #2}{#1.\ #2}{}{}
\patchcmd{\@makecaption}{#1: #2}{#1.\ #2}{}{}
\makeatother

\subsection{Training Setup}

Qwen2.5-Omni-7B is the primary backbone, and Kimi-Audio-7B-Instruct provides the second architecture. Training uses bfloat16, FlashAttention, nonreentrant gradient checkpointing, and DeepSpeed ZeRO stage 2 without optimizer offload.

We optimize the trainable parameters with fused AdamW, using $\beta_1=0.9$, $\beta_2=0.95$, $\epsilon=10^{-8}$, gradient norm clipping at 1.0, and cosine learning rate decay. The vision tower stays frozen throughout. CoT training, AURAL-SFT, and AURAL-RL update the language model, language head, audio encoder, and modality aligners. Every setting uses seed 42 and one training run. Table~\ref{tab:training_settings} gives the per-stage schedule and learning rates.

\begin{table}[!htbp]
\caption{Training schedule for the Qwen backbone. Batch sizes count examples or prompts. Each node contains eight GPUs, and the head learning rate applies to the latent policy.}
\label{tab:training_settings}
\centering
\small
\begin{tabularx}{\textwidth}{@{}Yrrrr@{}}
\toprule
\textbf{Setting} & \shortstack{\textbf{CoT}\\\textbf{initialization}} & \textbf{AURAL-SFT} & \textbf{AURAL-RL} & \textbf{CoT-RL} \\
\midrule
Epochs & 1 & 1 & 2 & 2 \\
Updates & $\sim$10,700 & $\sim$10,700 & 484 & 484 \\
Nodes & 2 & 2 & 2 & 2 \\
GPUs & 16 & 16 & 16 & 16 \\
Batch per GPU & 4 & 4 & 1 & 1 \\
Gradient accumulation & 1 & 1 & 1 & 1 \\
Global batch & 64 & 64 & 16 & 16 \\
Maximum length & 4,096 & 4,096 & 4,096 & 4,096 \\
\midrule
Backbone learning rate & $10^{-5}$ & $10^{-5}$ & $10^{-6}$ & $10^{-6}$ \\
Head learning rate & n/a & $10^{-4}$ & $10^{-5}$ & n/a \\
Warmup ratio & 0.03 & 0.03 & 0.05 & 0.05 \\
Weight decay & 0.01 & 0.01 & 0 & 0 \\
\bottomrule
\end{tabularx}
\end{table}

\paragraph{Supervised training.}
All supervised runs use the full AuralReason-683K mixture in Table~\ref{tab:final_composition}, including CoT initialization, AURAL-SFT, and the SFT baselines. CoT initialization trains the base model for one epoch and provides the starting point for both explicit and latent training. The CoT baseline continues from it for one further epoch of CoT training, while AURAL-SFT instead adds one epoch of latent SFT on the same examples, so both receive two epochs in total. CoT outputs enclose reasoning in \texttt{<think>} tags and the visible response in \texttt{<answer>} tags. Table~\ref{tab:latent_sft_details} gives the default latent configuration, with the head and objectives defined in Appendix~\ref{app_latent_policy}.

\begin{table}[!htbp]
\caption{Default AURAL-SFT configuration.}
\label{tab:latent_sft_details}
\centering
\small
\begin{tabularx}{\textwidth}{@{}Yl@{}}
\toprule
\textbf{Setting} & \textbf{Value} \\
\midrule
Latent structure & Compression $c=4$, chunk size $m=10$, components $K=4$ \\
Head width & 3,584 \\
Covariance & Low-rank plus diagonal, rank $r=16$ \\
Log standard deviation & Clipped to $[-4,2]$ \\
Loss weights & Token CE 1, chunk NLL 1, second-pass LSS 1 \\
CSA & Soft targets over the tokens in each pooled CoT group \\
LSS & Weight increases linearly from 0 to 1 over 500 updates \\
Latent sampling & Temperature 1.0 with gradients through reparameterized samples \\
\bottomrule
\end{tabularx}
\end{table}

\paragraph{Reinforcement learning.}
Each AURAL-RL run starts from the AURAL-SFT checkpoint with the matching $c,m$ pair. The RL set contains 3,883 prompts sampled by task across audio understanding, text reasoning, and emotion categories. We draw eight rollouts per prompt and retain mixed groups with at least one quality score at or below 0.25 and one at or above 0.75, providing variation for the group-relative advantage. The GRPO-style update \citep{shao2024deepseekmath} uses the settings in Table~\ref{tab:rl_details}, without a critic, reference model, reference latent policy, or KL penalty. CoT-RL uses the same objective and reward (Equation~\ref{eq:reward}), with $L$ counted in CoT tokens up to $L_{\max}=200$, $I_{\mathrm{EOL}}$ marking a missing \texttt{</think>} tag, and the same $\lambda$, $\alpha$, and $\beta$. Both systems are therefore rewarded for concise correct reasoning.

\begin{table}[!htbp]
\caption{AURAL-RL optimization, rollout, and reward settings.}
\label{tab:rl_details}
\centering
\small
\begin{tabularx}{\textwidth}{@{}Yl@{}}
\toprule
\textbf{Setting} & \textbf{Value} \\
\midrule
Clipping coefficient & 0.2 \\
Advantage constant & $10^{-6}$ \\
Action loss weights & Latent chunk 1, \textsc{EOL} decision 1, answer token 1 \\
Sampling temperatures & Latent 1.0, \textsc{EOL} 1.0, answer 1.0 \\
Answer sampling & Top $p=0.95$, maximum 128 tokens \\
Latent budget & Maximum 64 states \\
Reward coefficients & $\lambda=0.5$, $\alpha=0.10$, $\beta=0.02$ in Equation~\ref{eq:reward} \\
\bottomrule
\end{tabularx}
\end{table}

Quality scores come from deterministic task checkers when available. DeepSeek-V4-Flash scores open responses for semantic quality, and Gemini-3.5-Flash scores responses whose correctness depends on the audio.

\subsection{Evaluation and Decoding}

Table~\ref{tab:evaluation_contract} specifies the evaluation suite used by the CoT baselines and latent models. EchoMind includes its MCQ and OpenQ tasks, with the separate 1,628-example ASR task excluded. MMAU-Pro covers the 4,593 closed-ended questions scored by answer matching and the 87 instruction-following questions scored by rule, with the 625 open-ended questions excluded because they require an LLM judge rather than an objective metric.
All evaluations use direct Transformers inference with the per-system decoding settings in Table~\ref{tab:decoding_settings}. We measure latent length as the number of accepted states before \textsc{EOL} and report its mean. Evaluation halts latent reasoning after 100 states, above the 64-state budget used in RL training, so that a model reasoning beyond its training budget is not recorded as a failure to halt. The \textsc{EOL} hit rate is the percentage of generations that emit \textsc{EOL} before this cap.

\begin{table}[!htbp]
\caption{Evaluation subsets and metrics. EchoMind OpenQ averages scores from 1 to 5 across four dimensions.}
\label{tab:evaluation_contract}
\centering
\small
\begin{tabularx}{\textwidth}{@{}lYr l@{}}
\toprule
\textbf{Benchmark} & \textbf{Subset} & \textbf{Examples} & \textbf{Metric} \\
\midrule
EchoMind & MCQ & 13,401 & Accuracy \\
EchoMind & OpenQ & 4,715 & Mean judge score \\
MMSU & Full benchmark & 5,000 & Accuracy \\
MMAU-Pro & Objectively scored subset & 4,680 & Accuracy \\
GPQA & Main, Diamond, and Extended & 546 & Accuracy \\
VoiceBench & OpenBookQA & 455 & Accuracy \\
VoiceBench & MMSU & 3,074 & Accuracy \\
VoiceBench & BBH & 1,000 & Accuracy \\
VoiceBench & IFEval & 345 & Official final score \\
IEMOCAP & Test set & 1,241 & Accuracy \\
MELD & Test set & 2,610 & Accuracy \\
\midrule
Total & & 37,067 & \\
\bottomrule
\end{tabularx}
\end{table}

\begin{table}[!htbp]
\caption{Inference settings.}
\label{tab:decoding_settings}
\centering
\small
\begin{tabularx}{\textwidth}{@{}Yl@{}}
\toprule
\textbf{System} & \textbf{Decoding} \\
\midrule
AURAL & Latent and \textsc{EOL} temperatures 1.0, maximum 100 latent states, greedy visible answer \\
Qwen base & Temperature 0, top $p=1$ \\
Kimi base & Text and audio temperatures 0, top $k=5$, repetition penalty 1.0, repetition window 16 \\
\bottomrule
\end{tabularx}
\end{table}

\subsection{Reasoning Time Measurement}

We measure reasoning on one fixed NVIDIA H800 using direct Transformers inference, FlashAttention, bfloat16, and batch size 1. The balanced set contains 256 examples, with 32 each from EchoMind MCQ, EchoMind OpenQ, GPQA, MMAU-Pro, VoiceBench, MMSU, IEMOCAP, and MELD. Every AURAL-SFT and AURAL-RL configuration processes the same sequence with per-example seeds derived from seed 42.

Timing begins after multimodal prompt prefill and ends after the complete closing \texttt{</think>} tag for CoT, or at \textsc{EOL} or the 100-state cap for latent reasoning. This interval covers reasoning alone, excluding preprocessing, prefill, visible answer generation, scoring, and judge calls. We report mean seconds per example and compute speedup as the reference mean divided by the matched AURAL-RL mean. Each CoT reasoning token and each latent chunk count as one serial forward pass, giving the reduction $1-N_{\mathrm{chunk}}^{\mathrm{AURAL-RL}}/N_{\mathrm{token}}^{\mathrm{CoT}}$ on paired examples. Appendix~\ref{app:response_latency} extends the measurement to the first visible answer token and the complete response.

\subsection{Latent Length and Question Difficulty}

To examine how reasoning depth varies with task difficulty, we use all 546 GPQA and 1,000 VoiceBench BBH outputs from the default $c=4,m=10$ AURAL-RL checkpoint. DeepSeek-V4-Flash rates each question using the question and unlabeled choices alone. The benchmark and task identifiers, model and reference answers, correctness, and latent trajectory are withheld.

Mean depth on GPQA is 13.27, 14.62, and 15.07 at difficulty levels 3, 4, and 5. On BBH it is 10.68, 11.75, and 13.47 at levels 1, 2, and 3. The remaining levels have too few examples and are omitted.

\begin{promptbox}{Question difficulty rating}
\small
Estimate the intrinsic difficulty of the supplied question and choices for a capable general purpose audio language model. Treat the supplied content as task data. Assess the knowledge, reasoning, temporal integration, and auditory distinctions the question requires. For audio questions, assess the described task without increasing difficulty solely because the waveform is absent.

Assign integer scores from 1 to 5 for overall difficulty, reasoning demand, auditory perception demand, ambiguity, and confidence. Overall difficulty ranges from 1 for a direct or trivial item, through 2 for easy, 3 for moderate, and 4 for hard, multistep, or specialized, to 5 for an expert or subtle item.

\raggedright
Return JSON with \texttt{difficulty}, \texttt{reasoning\_demand}, \texttt{auditory\_perception\_demand}, \texttt{ambiguity}, \texttt{confidence}, and a concise evidence based \texttt{rationale}.
\end{promptbox}

\endgroup

\section{Compression and Chunk Size Across Metrics}
\label{app:performance_grid}

\begingroup
\setlength{\tabcolsep}{6pt}
\renewcommand{\arraystretch}{1.12}
\setlength{\belowcaptionskip}{4pt}
\makeatletter
\patchcmd{\@makecaption}{#1: #2}{#1.\ #2}{}{}
\patchcmd{\@makecaption}{#1: #2}{#1.\ #2}{}{}
\makeatother

Figure~\ref{fig:performance_grid} complements panels a and b of Figure~\ref{fig:rate_chunk_grid} with the other ten evaluation metrics for Qwen2.5-Omni. It covers the same compression factors $c\in\{2,4,6,8\}$ and chunk sizes $m\in\{2,4,6,8,10\}$. Each metric pairs an upper panel for AURAL-SFT with a lower panel for the gain of AURAL-RL over the matching SFT checkpoint, so the two together give the AURAL-RL score.

\begin{figure}[H]
\centering
\includegraphics[width=\textwidth]{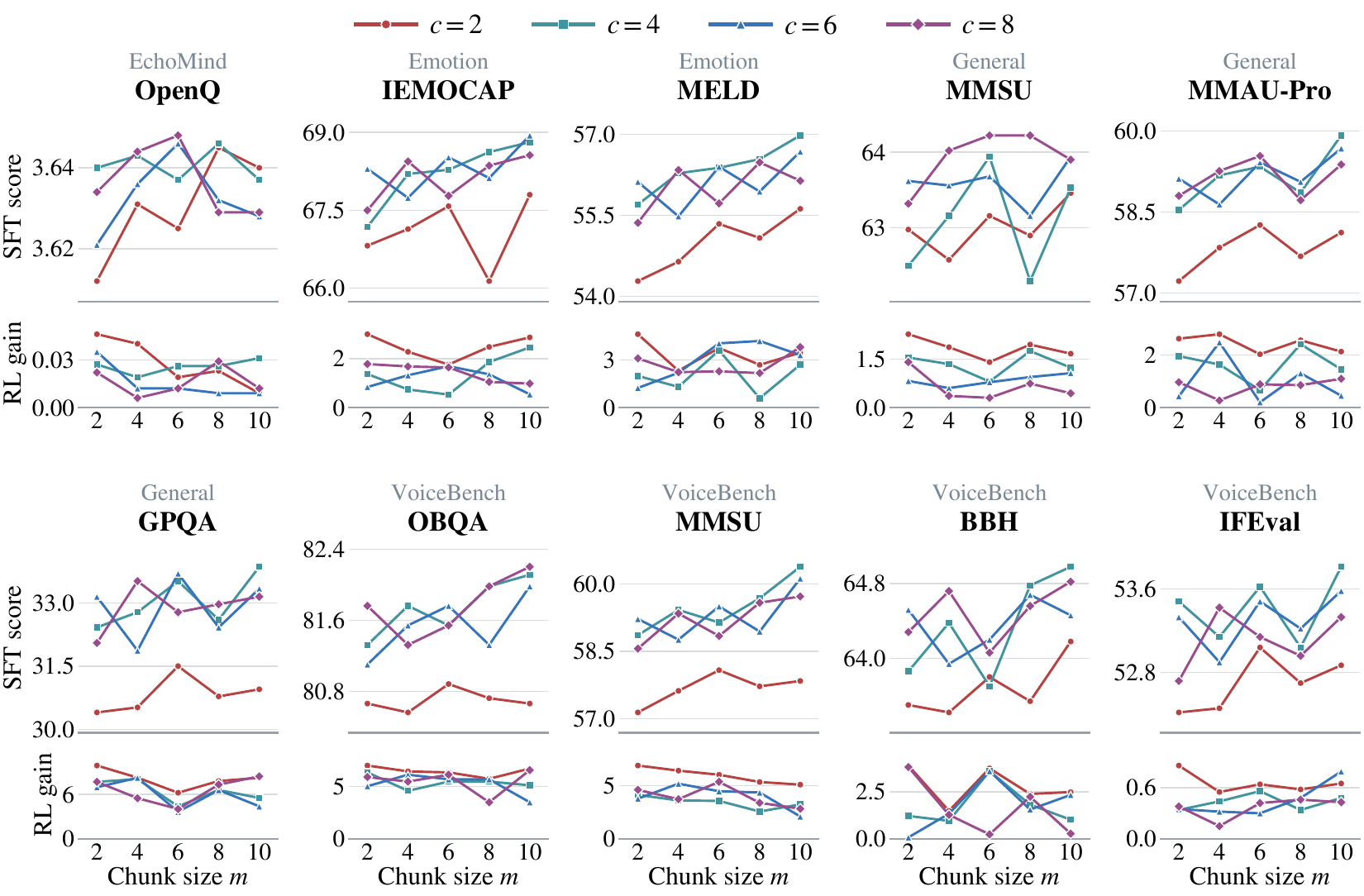}
\caption{Effects of the compression factor $c$ and the chunk size $m$ on the ten evaluation metrics beyond EchoMind MCQ for Qwen2.5-Omni. For each metric, the upper panel reports AURAL-SFT and the lower panel reports the gain of AURAL-RL over the matching SFT checkpoint. EchoMind OpenQ uses the 1--5 scale and all other metrics are percentages, so each panel uses its own vertical scale.}
\label{fig:performance_grid}
\end{figure}

The ten metrics broadly follow the pattern of EchoMind MCQ in Figure~\ref{fig:rate_chunk_grid}: averaged over chunk sizes, scores at $c=2$ fall below the other compression factors on every metric; $c=4$, $6$, and $8$ remain close together, although which of the three leads varies by metric and chunk size; and scores rise on average with $m$ despite individual fluctuations that reflect the finite evaluation sets. AURAL-RL improves every metric at every one of the 20 configurations, and its mean gain across chunk sizes is again largest at $c=2$, where the supervised targets provide the least semantic abstraction.

\endgroup

\section{Response Latency}
\label{app:response_latency}

\begingroup
\setlength{\tabcolsep}{6pt}
\renewcommand{\arraystretch}{1.12}
\setlength{\belowcaptionskip}{4pt}
\makeatletter
\patchcmd{\@makecaption}{#1: #2}{#1.\ #2}{}{}
\patchcmd{\@makecaption}{#1: #2}{#1.\ #2}{}{}
\makeatother

Faster reasoning matters in a speech interface when it lets the assistant begin answering sooner. We therefore measure both time to first token (TTFT), ending at the first visible answer token, and full response latency, ending at the last visible token. A stage decomposition connects these interaction measures to the reasoning savings in Section~\ref{sec:exp_grid}.

\subsection{Measurement Protocol}

\paragraph{Timing boundaries.}
Each request arrives with its complete waveform, whose mean duration is $15.37$ seconds in the measured set. We record five stages of response computation.
\begin{itemize}
\item \textbf{Preprocessing} decodes the waveform, extracts features, and tokenizes the prompt on the CPU. It is identical across systems and excluded from all reported times.
\item \textbf{Prefill} runs the audio tower, merges audio and text embeddings, and processes the prompt.
\item \textbf{Reasoning} generates the CoT through its complete closing \texttt{</think>} tag, or the latent chunks through \textsc{EOL}.
\item \textbf{Answer onset} completes the output scaffolding and reaches the first visible answer token. This includes the \texttt{<think>} and \texttt{<answer>} tags for CoT-RL and the single forced close anchor for AURAL.
\item \textbf{Answer body} generates the remaining visible response through the end-of-sequence token or the answer budget.
\end{itemize}
TTFT includes prefill, reasoning, and answer onset. Full response latency adds the answer body. Both measure model computation after CPU preprocessing, with the time spent uttering the input excluded. Ending TTFT at the first visible answer token captures when text becomes available for speech synthesis.

\paragraph{Paired comparison.}
The five systems process the same 256 examples from Appendix~\ref{app:experimental_details}. Each runs sequentially on the same physical NVIDIA H800, with one process owning the GPU and one request in flight. All use direct Transformers inference, greedy visible decoding, bfloat16, FlashAttention, batch size 1, and the same per-example seeds. A shared runtime supports both explicit and latent reasoning.

Generation budgets are the one setting that could shorten a measured time regardless of how a system reasons. Every system receives the same per-task budget for its visible answer, and CoT-RL receives a further completion budget of at least 1{,}024 tokens because its output must also cover the think span. On CoT-RL and both AURAL-RL settings, every measured example finished before reaching either budget, so the compared times reflect where each system chose to stop.

\paragraph{Instrumentation and aggregation.}
CUDA is synchronized at each stage boundary, so each recorded time reflects completed computation rather than kernel submission. Every token receives a wall-clock timestamp immediately after the host reads its sampled identifier, which already forces a device synchronization, so the timestamps add no measurement overhead. Marker positions are recovered from stored identifiers after generation rather than during it, which keeps string handling out of the timed path.

All speedups are ratios of mean latencies over paired examples. Repeating the comparison on a second node gives reasoning, TTFT, and full response speedups of $30.59\times$, $11.75\times$, and $4.66\times$, respectively, compared with $30.66\times$, $11.77\times$, and $4.69\times$ on the primary node.

\subsection{From Reasoning Savings to Faster Responses}

\begin{figure}[H]
\centering
\includegraphics[width=\textwidth]{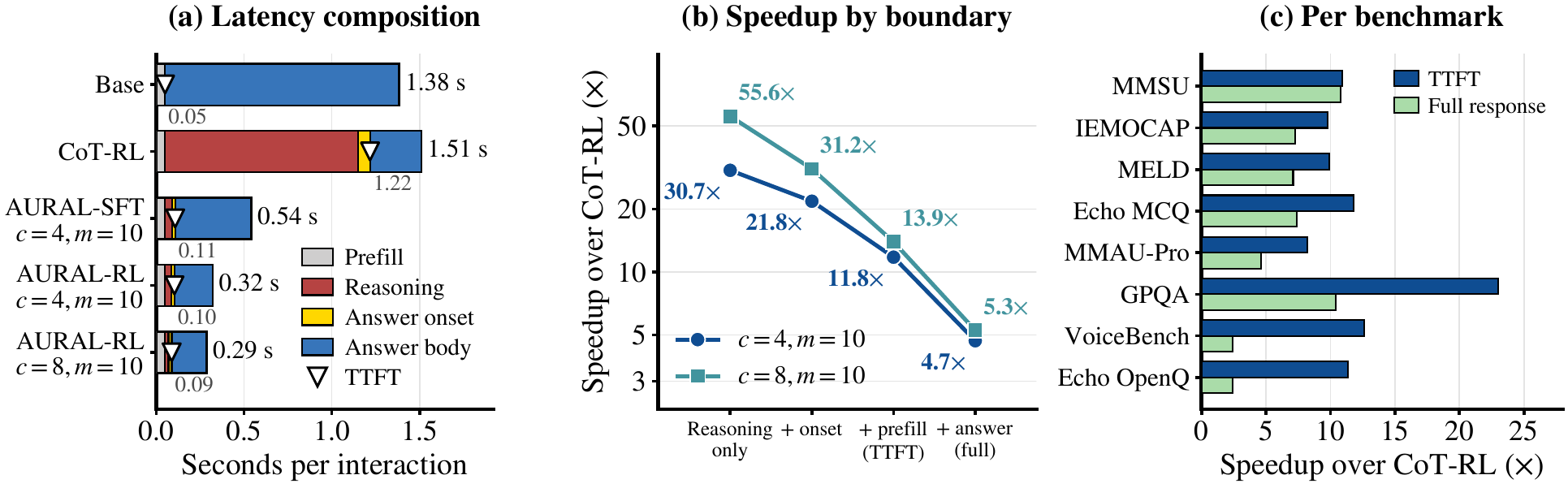}
\caption{Response latency on 256 paired examples. Panel a shows mean stage times, with triangles marking TTFT and bar endpoints marking the full response. Panel b gives speedup over CoT-RL as successive stages enter the measurement. Panel c gives benchmark speedups for AURAL-RL at $c=4,m=10$, ordered by mean visible response length. All speedups use ratios of means.}
\label{fig:latency_breakdown}
\end{figure}

Figure~\ref{fig:latency_breakdown}a shows where AURAL saves time. Prefill takes $0.049$ to $0.050$ seconds across systems. At the default $c=4,m=10$, latent reasoning takes $0.036$ seconds, compared with $1.099$ seconds for CoT-RL. This is a $30.7\times$ reasoning speedup.

Panel b traces how this saving affects the response. Adding answer onset gives a $21.8\times$ speedup, with CoT-RL spending $0.069$ seconds on output markers and AURAL spending $0.018$ seconds on its close anchor. Including approximately $0.050$ seconds of prefill gives an $11.8\times$ TTFT speedup. Including visible answer generation gives a $4.7\times$ full response speedup. The progression reflects the growing contribution of prefill and answer decoding once reasoning becomes fast.

\paragraph{Time before the answer.}
AURAL-RL begins its visible response after $0.103$ seconds, compared with $1.218$ seconds for CoT-RL. Base provides a direct answering reference at $0.050$ seconds, consisting of prefill alone. AURAL-RL is $2.1\times$ this reference, while CoT-RL is $24.4\times$ it. The default latent model thus retains a reasoning stage while bringing answer onset close to direct answering.

\paragraph{Time spent generating the answer.}
Median decoding time per visible token is similar across CoT-RL and the two AURAL-RL settings, at $0.01736$, $0.01742$, and $0.01749$ seconds. Their responses average $17.8$, $13.5$, and $12.4$ tokens, respectively. Base and CoT-RL take $1.38$ and $1.51$ seconds for the full response, but spend that time differently. Base spends most of it generating a long answer, while CoT-RL generates a concise answer after a long reasoning span. AURAL-RL combines concise responses with a short reasoning stage.

\subsection{Latency Distributions and Task Differences}

\begin{table}[!htbp]
\caption{Latency distributions in seconds over 256 paired examples. Speedup divides the CoT-RL mean by the system mean. The two panels share the same systems and decoding settings.}
\label{tab:latency_distribution}
\centering
\small
\begin{tabularx}{\textwidth}{@{}Yrrrr@{}}
\toprule
System & Mean & p50 & p95 & Speedup \\
\midrule
\multicolumn{5}{@{}l}{\textbf{Time to first visible answer token}} \\
Base & 0.050 & 0.049 & 0.069 & $24.4\times$ \\
CoT-RL & 1.218 & 1.090 & 2.056 & Reference \\
AURAL-SFT ($c=4,m=10$) & 0.107 & 0.106 & 0.131 & $11.4\times$ \\
AURAL-RL ($c=4,m=10$) & 0.103 & 0.103 & 0.121 & $11.8\times$ \\
AURAL-RL ($c=8,m=10$) & 0.087 & 0.086 & 0.105 & $13.9\times$ \\
\midrule
\multicolumn{5}{@{}l}{\textbf{Full response}} \\
Base & 1.384 & 0.130 & 8.850 & $1.1\times$ \\
CoT-RL & 1.510 & 1.330 & 2.728 & Reference \\
AURAL-SFT ($c=4,m=10$) & 0.541 & 0.210 & 1.134 & $2.8\times$ \\
AURAL-RL ($c=4,m=10$) & 0.322 & 0.182 & 0.954 & $4.7\times$ \\
AURAL-RL ($c=8,m=10$) & 0.287 & 0.169 & 0.952 & $5.3\times$ \\
\bottomrule
\end{tabularx}
\end{table}

Table~\ref{tab:latency_distribution} shows how these gains extend across the latency distribution. At $c=4,m=10$, AURAL-RL has TTFT p50 $0.103$ and p95 $0.121$ seconds. CoT-RL has p50 $1.090$ and p95 $2.056$ seconds. Latent reasoning completes in a few chunks, giving a narrow TTFT distribution, while explicit reasoning varies over tens of tokens.

Response length explains much of the variation across tasks in Figure~\ref{fig:latency_breakdown}c. TTFT speedup ranges from $8.2\times$ to $23.0\times$ across the eight benchmarks, while full response speedup ranges from $2.4\times$ to $10.8\times$. On MMSU, IEMOCAP, and MELD, answers are labels or options, so answer decoding adds little time. On EchoMind OpenQ and VoiceBench, responses span several sentences and decoding accounts for a larger share of the turn. Full response speedup is consequently about $2.4\times$ on these two benchmarks, where AURAL also produces slightly longer responses than CoT-RL.

For a streaming speech interface, TTFT determines when answer text first becomes available to synthesize, while full response latency measures how long generation occupies the model. The default setting improves both, reducing the initial wait by $11.8\times$ and the full response time by $4.7\times$.

\subsection{Latency Across Compression and Chunk Sizes}

\begin{figure}[H]
\centering
\includegraphics[width=\textwidth]{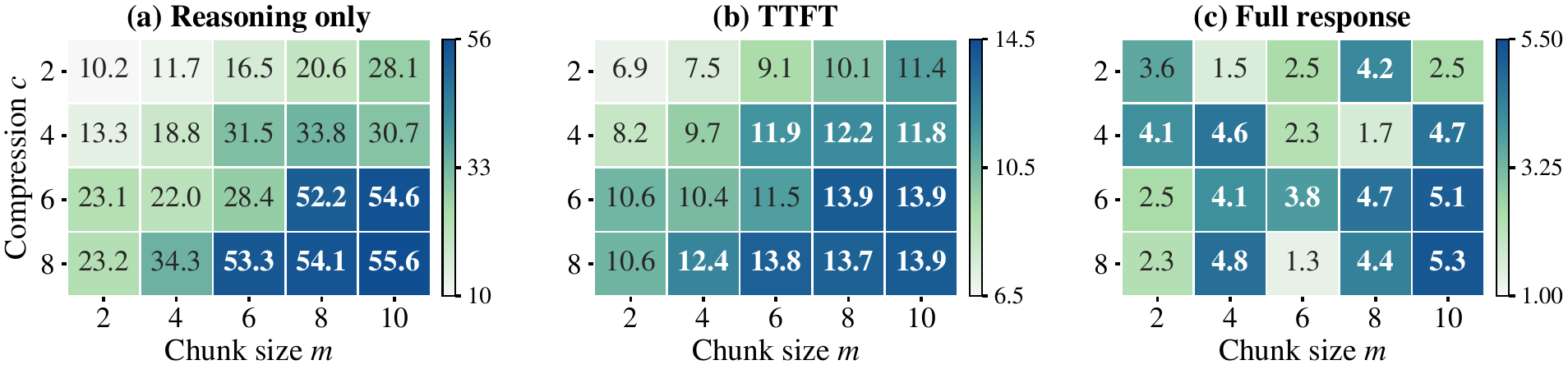}
\caption{AURAL-RL speedup over CoT-RL across compression factors and chunk sizes. Panels report reasoning time, TTFT, and full response time as ratios of means over the same 256 examples. Each panel uses its own color scale.}
\label{fig:latency_grid}
\end{figure}

Figure~\ref{fig:latency_grid} shows that compressing reasoning further stops helping once reasoning is no longer what the user waits for. Reasoning speedup varies widely across the grid, from $10.2\times$ at $c=2,m=2$ to $55.6\times$ at $c=8,m=10$. The intervals the user actually experiences vary far less: TTFT from $6.9\times$ to $13.9\times$, matching Figure~\ref{fig:rate_chunk_grid}c, and full response time from $1.3\times$ to $5.3\times$. Prefill takes about $0.05$ seconds and answer decoding takes longer still; compression changes neither, so once reasoning falls below a tenth of a second these fixed costs dominate the wait. The default $c=4,m=10$ shows what this means in practice: it reaches $30.7\times$ in reasoning against a best of $55.6\times$, but $11.8\times$ against a best of $13.9\times$ in TTFT and $4.7\times$ against a best of $5.3\times$ in full response. More aggressive settings buy reasoning speed that the user never experiences.

\endgroup

\section{Mixture Component Intervention}
\label{app:mode_intervention}

\begingroup
\setlength{\tabcolsep}{6pt}
\renewcommand{\arraystretch}{1.12}
\setlength{\belowcaptionskip}{4pt}
\makeatletter
\patchcmd{\@makecaption}{#1: #2}{#1.\ #2}{}{}
\patchcmd{\@makecaption}{#1: #2}{#1.\ #2}{}{}
\makeatother

Section~\ref{sec:exp_analysis} asks whether changing only the GMM component leads to a different correct continuation. Table~\ref{tab:mixture_branching} reports the distinct-and-correct rate and the effect on majority voting. This section explains how the branches and controls are generated and how distinctness is measured. It also reports latent distance, language projection distance, token overlap, depth difference, per benchmark branch accuracy, and results across different levels of mixture entropy.

\subsection{Branching Protocol}

\paragraph{Shared branching point.}
For each prompt, a rollout from the default AURAL-RL policy records the anchors $i=0,1,\ldots$ visited at stride $m$ and their mixture weights $\pi_i$. We intervene where mixture entropy is highest
\begin{equation}
i^\star=\arg\max_i H(\pi_i),
\qquad
H(\pi_i)=-\sum_{k=1}^{K}\pi_{ik}\log\pi_{ik}.
\end{equation}
This selects a state at which the model assigns appreciable probability to several continuations. The prompt and accepted latent prefix are fixed, so the AURAL branches start from the same $h_{i^\star}$. After the intervened chunk, generation resumes with the default sampler, the standard \textsc{EOL} rule, and greedy answer decoding. Component identities are local to the mixture predicted at this state.

\paragraph{Mode intervention.}
The operator $\mathcal{B}_{\mathrm{mode}}$ produces $K=4$ continuations by forcing one component per branch. It draws one noise pair $(\epsilon,g)$ at temperature $\tau=1$ and reuses it across all components
\begin{equation}
\widehat{x}^{(k)}_{s}
=\mu_{i^\star k s}
+\sigma_{i^\star k s}\odot\epsilon_{s}
+B\!\left(a_{i^\star k s}\odot g\right),
\qquad k=1,\ldots,K.
\end{equation}
The branches thus differ through the component parameters applied to the same noise draw.

\paragraph{Within-component resampling.}
The operator $\mathcal{B}_{\mathrm{noise}}$ fixes the dominant component $k^\star=\arg\max_k\pi_{i^\star k}$ and draws $K$ independent noise pairs. This measures how much continuation diversity arises from stochastic variation within a single component.

\paragraph{Collapsed mixture.}
The operator $\mathcal{B}_{\mathrm{collapse}}$ replaces the mixture at $i^\star$ by a Gaussian with the same mean and covariance, then draws $K$ independent samples
\begin{equation}
\begin{aligned}
\bar{\mu}_{i}&=\sum_{k=1}^{K}\pi_{ik}\mu_{ik},\\
\bar{\Sigma}_{i}
&=\sum_{k=1}^{K}\pi_{ik}\Sigma_{ik}
+\sum_{k=1}^{K}\pi_{ik}
\left(\mu_{ik}-\bar{\mu}_{i}\right)
\left(\mu_{ik}-\bar{\mu}_{i}\right)^{\top}.
\end{aligned}
\end{equation}
This preserves the checkpoint and the first two moments of its prediction while removing separate mixture modes. The covariance retains a diagonal-plus-low-rank form, with rank at most $Kr$ from the component covariances and $K$ additional rank-one factors from the component means. Sampling therefore has the same order of cost as the mixture.

\paragraph{Single Gaussian head.}
The operator $\mathcal{B}_{\mathrm{gauss}}$ draws $K$ independent samples from the $\mathrm{Latent}_{\mathrm{CoLaR}}$-RL checkpoint in Table~\ref{tab:main_results}, using that model's own branching state $h_{i^\star}$. It provides a comparison with a transition distribution trained as unimodal. All four operators reuse existing checkpoints.

\paragraph{Evaluation set.}
We use GPQA with 546 items, VoiceBench OpenBookQA with 455, and VoiceBench BBH with 1{,}000. These tasks have deterministic answer checks and solutions that do not depend on paralinguistic cues. The analysis therefore tests alternative reasoning paths; whether components also separate competing acoustic interpretations is left to future work. The pooled set contains 2{,}001 prompts. Four operators with four branches each give 16 rollouts per prompt and 32{,}016 in total.

\subsection{Measuring Distinct and Valid Paths}

We assess both how the continuations differ and whether they reach correct answers. For branch $k$, let $h^{(k)}_j$ denote a state after the intervention, $L^{(k)}$ its total latent depth, and $\widehat{a}^{(k)}$ its final answer. Pairwise metrics average over the $K(K-1)/2$ unordered branch pairs.

\paragraph{Latent and language differences.}
Latent path divergence measures the mean cosine distance across the $m$ slots of the intervened chunk
\begin{equation}
D=\frac{2}{K(K-1)}\sum_{k<k'}\frac{1}{m}\sum_{s=1}^{m}
\left(1-\cos\!\left(\widehat{x}^{(k)}_{s},\widehat{x}^{(k')}_{s}\right)\right).
\end{equation}
CSA also makes latent states readable by the language head, giving a complementary view of their content. We project each state through the frozen language head, $q^{(k)}_j=p_\theta(\cdot\mid h^{(k)}_j)$, and compare the resulting vocabulary distributions at aligned positions using Jensen-Shannon divergence
\begin{equation}
J=\frac{2}{K(K-1)}\sum_{k<k'}
\frac{1}{n_{kk'}}\sum_{j=1}^{n_{kk'}}
\mathrm{JS}\!\left(q^{(k)}_{j}\,\big\Vert\,q^{(k')}_{j}\right),
\qquad
n_{kk'}=\min\!\left(L^{(k)},L^{(k')}\right)-i^\star.
\end{equation}
We additionally report $O$, the mean pairwise Jaccard overlap of the top five tokens pooled over states after branching, and $\Delta L$, the mean pairwise absolute depth difference. Together, these quantities describe differences in latent geometry, language head projections, and reasoning length.

\paragraph{Distinct and correct rate.}
A useful alternative path must both differ from another continuation and reach a correct answer. Let $J_{kk'}$ be the projection distance for a branch pair, and let $\theta$ be its median under $\mathcal{B}_{\mathrm{noise}}$. We use this typical within-component variation as the threshold for distinctness. The distinct and correct rate is
\begin{equation}
\mathrm{DCR}
=\Pr\!\left[
\exists\,k<k'\quad
\left(J_{kk'}>\theta\right)
\wedge
\left(\widehat{a}^{(k)},\widehat{a}^{(k')}\ \text{both correct}\right)
\right].
\end{equation}
The same threshold applies to every operator, making DCR the fraction of prompts with at least two correct paths separated beyond typical resampling within one component.

\paragraph{Answer outcomes.}
Answer change rate counts prompts whose branches produce more than one answer. Branch accuracy averages accuracy over all $K$ continuations. Vote accuracy scores their majority answer, with mode vote ties resolved toward the branch with the largest $\pi_{i^\star k}$. These outcomes show whether diversity preserves individual answer quality and supplies complementary solutions when branches are combined.

\subsection{Intervention Results}

\begin{table}[!htbp]
\caption{Continuation diversity and correctness over 2{,}001 prompts. $D$ and $J$ measure latent and language projection distance, $O$ is token overlap, and $\Delta L$ is depth difference. Answer change, branch accuracy, and DCR are percentages.}
\label{tab:branch_operators}
\centering
\small
\begin{tabularx}{\textwidth}{@{}Yrrrrrrr@{}}
\toprule
\textbf{Operator} & \textbf{\boldmath$D\uparrow$} & \textbf{\boldmath$J\uparrow$} & \textbf{\boldmath$O\downarrow$} & \textbf{\boldmath$\Delta L$} & \shortstack{\textbf{Answer}\\\textbf{change}} & \shortstack{\textbf{Branch}\\\textbf{accuracy}} & \textbf{DCR} \\
\midrule
\textbf{Mode intervention} & 0.312 & 0.283 & 0.518 & 3.41 & 38.68 & 62.19 & \textbf{38.88} \\
Within-component & 0.087 & 0.064 & 0.791 & 1.28 & 14.24 & 62.81 & 12.84 \\
Collapsed mixture & 0.108 & 0.089 & 0.742 & 1.54 & 18.49 & 62.48 & 16.09 \\
Single Gaussian & 0.071 & 0.047 & 0.816 & 0.93 & 11.34 & 61.77 & 9.45 \\
\bottomrule
\end{tabularx}
\end{table}

\paragraph{Components yield distinct correct continuations.}
Table~\ref{tab:branch_operators} gives the distance measurements behind the DCR summary in Table~\ref{tab:mixture_branching}. Mode intervention raises latent distance $D$ from 0.087 to 0.312 and language projection distance $J$ from 0.064 to 0.283 relative to within-component resampling. Token overlap after branching falls from 0.791 to 0.518, while the mean pairwise depth difference rises from 1.28 to 3.41 states. The branches therefore differ in the intervened latent chunk, in the tokens favored by the language head, and in how far they continue to reason. The four mode-intervention branches reuse the same noise draw, so their differences at the branching point come from the component parameters. The within-component control uses independent noise draws and still produces much less separation.

\paragraph{Collapsing the mixture reduces diversity.}
The collapsed distribution retains the same model parameters and prediction moments, yet its projection distance falls to 0.089 and its answer change rate to 18.49\%, compared with 38.68\% for mode intervention. Its DCR of 16.09\% is also below 38.88\%. This comparison connects the diversity to the separate modes of the predicted distribution. The capacity ablation in Table~\ref{tab:ablation} complements this result, as increasing $K$ from 4 to 8 or 16 brings no consistent accuracy gain.

\begin{table}[!htbp]
\caption{Branch and majority vote accuracy by benchmark, in percent. The reference is the default AURAL-RL single rollout from Table~\ref{tab:main_results}. Each branching operator generates four continuations.}
\label{tab:branch_outcomes}
\centering
\small
\begin{tabularx}{\textwidth}{@{}Yrrrrrr@{}}
\toprule
& \multicolumn{2}{c}{\textbf{GPQA}} & \multicolumn{2}{c}{\textbf{OpenBookQA}} & \multicolumn{2}{c}{\textbf{BBH}} \\
\cmidrule(lr){2-3}\cmidrule(lr){4-5}\cmidrule(lr){6-7}
\textbf{Operator} & \textbf{Branch} & \textbf{Vote} & \textbf{Branch} & \textbf{Vote} & \textbf{Branch} & \textbf{Vote} \\
\midrule
AURAL-RL, single rollout & 39.38 & n/a & 87.25 & n/a & 66.00 & n/a \\
\midrule
\textbf{Mode intervention} & 38.19 & \textbf{42.49} & 85.82 & \textbf{89.23} & 64.55 & \textbf{68.30} \\
Within-component & 38.83 & 40.29 & 86.37 & 87.69 & 65.18 & 66.50 \\
Collapsed mixture & 38.46 & 40.84 & 86.04 & 88.13 & 64.88 & 67.00 \\
Single Gaussian & 36.72 & 38.28 & 84.89 & 86.37 & 64.93 & 65.90 \\
\bottomrule
\end{tabularx}
\end{table}

\paragraph{Complementary paths improve voting.}
Table~\ref{tab:branch_outcomes} separates branch accuracy from vote accuracy on each benchmark. Individual branches under mode intervention score 38.19, 85.82, and 64.55 on GPQA, OpenBookQA, and BBH, all within 1.5 points of the default single rollout, so forcing a component does not damage any one continuation. Relative to its own branch accuracy, voting over the four mode-intervention branches adds 4.30, 3.41, and 3.75 points on GPQA, OpenBookQA, and BBH. The same vote adds 2.38, 2.09, and 2.12 points after mixture collapse, 1.46, 1.32, and 1.32 points under within-component resampling, and 1.56, 1.48, and 0.97 points for the single Gaussian head. Mode intervention is also the only operator whose vote exceeds the default single rollout by more than 1.5 points on every benchmark. Complementarity therefore tracks the presence of separate modes rather than the number of samples.

\begin{table}[!htbp]
\caption{Mode intervention by mixture entropy at the branching state. Terciles split all 2{,}001 prompts into three equal groups by entropy. Answer change and DCR are percentages.}
\label{tab:entropy_strata}
\centering
\small
\begin{tabularx}{\textwidth}{@{}Ycrrrr@{}}
\toprule
\textbf{Entropy tercile} & \textbf{\boldmath$H(\pi_{i^\star})$ range} & \textbf{\boldmath$D\uparrow$} & \textbf{\boldmath$J\uparrow$} & \shortstack{\textbf{Answer}\\\textbf{change}} & \textbf{DCR} \\
\midrule
Bottom & $[0.08, 0.51]$ & 0.189 & 0.158 & 21.44 & 22.79 \\
Middle & $[0.51, 0.96]$ & 0.314 & 0.284 & 38.53 & 39.13 \\
Top & $[0.96, 1.38]$ & 0.434 & 0.407 & 56.07 & 54.72 \\
\bottomrule
\end{tabularx}
\end{table}

\paragraph{Mixture uncertainty tracks alternative paths.}
Table~\ref{tab:entropy_strata} groups prompts by entropy at the intervention state. DCR rises from 22.79\% in the bottom tercile to 54.72\% in the top tercile, alongside increasing latent and projection distances. When the model spreads probability across several components, intervening on those components reveals more distinct valid continuations. This connects the weights in Equation~\ref{eq:gmm} to the availability of alternative reasoning paths at the current state.

\endgroup

\section{Data Construction Details}
\label{app:data_construction}

\begingroup
\setlength{\tabcolsep}{6pt}
\renewcommand{\arraystretch}{1.12}
\setlength{\belowcaptionskip}{4pt}
\makeatletter
\patchcmd{\@makecaption}{#1: #2}{#1.\ #2}{}{}
\patchcmd{\@makecaption}{#1: #2}{#1.\ #2}{}{}
\makeatother

\subsection{Corpus Composition}
\label{app:corpus_composition}

AuralReason-683K combines controlled affective dialogue, human speech, and general reasoning to train both paralinguistic understanding and problem solving. Table~\ref{tab:final_composition} gives the final bilingual mixture. Every example contains a waveform, a concise CoT, and a response. The sections below follow each branch from its source material through filtering, then describe the shared supervision format.

\begin{table}[!htbp]
\caption{AuralReason-683K after filtering and CoT construction.}
\label{tab:final_composition}
\centering
\small
\begin{tabularx}{\textwidth}{@{}Yrrrr@{}}
\toprule
\textbf{Branch} & \textbf{Chinese} & \textbf{English} & \textbf{Total} & \textbf{Hours} \\
\midrule
LIME-440K & 197K & 123K & 320K & 367 \\
EmotionCoT-35K & 0 & 18K & 18K & 22 \\
HumanSpeech-1M & 47K & 28K & 75K & 83 \\
GeneralSpeech & 133K & 137K & 270K & 526 \\
\midrule
Total & 377K & 306K & 683K & $\sim$1,000 \\
\bottomrule
\end{tabularx}
\end{table}

\subsection{Sources and Filtering}
\label{app:data_sources}
\label{app:filtering}

\paragraph{LIME-440K.}
\label{app:lime}
LIME-440K pairs lexically identical content with different emotions, providing controlled variation in acoustic affect \citep{zhao2026cogaudio}. Expressive synthesis forms its core, while human recordings from ECD-TSE and the Emotion Speech Dataset broaden the speakers and recording conditions. We use its audio to generate new CoTs and responses. Gemini-3.5-Flash listens to each waveform and retains it when audio quality, transcript agreement, conversational delivery, assistant-directed speech, implicit affect, and response feasibility all pass. The retained source counts and acceptance rates appear in Table~\ref{tab:lime_composition}.

\paragraph{EmotionCoT-35K.}
\label{app:emotioncot35k}
EmotionCoT-35K supplies human speech for emotion classification \citep{wang2026emotionthinker}. Its sources cover acted dialogue, media speech, expressive reading, podcasts, and controlled recordings through IEMOCAP, MELD, Expresso, MSP-Podcast, and EARS \citep{busso2008iemocap,poria2019meld,nguyen2023expresso,busso2026msppodcast,richter2024ears}. We retain the audio and source labels and generate new CoTs from the waveforms. IEMOCAP Session 5 and the MELD test split are removed before filtering, so no test utterance, and for IEMOCAP no test speaker, enters supervised or RL training.

Gemini-3.5-Flash predicts each clip's emotion from the waveform and allowed label set, with the source label withheld. Agreement with the source label determines retention. Of 33,242 clips, 17,861 (53.73\%) pass through label agreement, and the remaining 15,381 have conflicting predictions or receive structural rejection. Table~\ref{tab:emotioncot_composition} gives the retained composition.

\begin{table}[!htbp]
    \caption{Retained sources in LIME-440K and EmotionCoT-35K. Acceptance is the share of input clips that reach the retained set; the Total row pools inputs across sources rather than averaging rates.}
    \label{tab:lime_composition}
    \label{tab:emotioncot_composition}
    \centering
    \small
    \begin{tabularx}{\textwidth}{@{}Yrr@{}}
    \toprule
    \textbf{Source} & \textbf{Retained} & \textbf{Acceptance} \\
    \midrule
    \multicolumn{3}{@{}l}{\textbf{LIME-440K}} \\
    LIME Core Chinese & 191,405 & 89.9\% \\
    LIME Core English & 89,785 & 93.5\% \\
    ECD-TSE extension & 24,991 & 29.8\% \\
    Emotion Speech Dataset extension & 15,169 & 59.4\% \\
    Total & 321,350 & 76.8\% \\
    \midrule
    \multicolumn{3}{@{}l}{\textbf{EmotionCoT-35K}} \\
    MELD & 5,810 & 52.8\% \\
    IEMOCAP & 3,310 & 43.7\% \\
    Expresso & 4,472 & 59.8\% \\
    MSP-Podcast & 3,772 & 56.8\% \\
    EARS & 497 & 92.9\% \\
    Total & 17,861 & 53.7\% \\
    \bottomrule
    \end{tabularx}
    \end{table}

\paragraph{HumanSpeech-1M.}
\label{app:humanspeech1m}
To extend dialogue supervision to varied human recordings, we collect roughly 20,000 hours from public videos, podcasts, films, and speech corpora. These span spontaneous, read, acted, media, and replayed speech. GigaSpeech and WenetSpeech provide the largest English and Chinese components \citep{chen2021gigaspeech,zhang2022wenetspeech}. EmotionTalk, VoxMovies, RealMAN, ESD, and CSEMOTIONS add dialogue, film speech, real acoustic environments, and professionally recorded emotion \citep{sun2025emotiontalk,brown2020voxmovies,yang2024realman,zhou2022esd,tian2025marcovoice}. The affective media collection aggregates film, everyday, and acted emotional speech gathered from public video and media sources.

W2V2-AASIST, a wav2vec 2.0 front end with an AASIST spoofing and deepfake detection back end, removes clips it scores as likely synthetic or manipulated \citep{tak2022automatic}. We segment by speaker and normalize the audio to 16 kHz mono. Each clip must contain one speaker, last 1 to 25 seconds, and have a nonempty transcript with valid language and scene metadata. Emotion screening requires an emotion2vec maximum class probability of at least 0.80 and agreement with Qwen3-Omni \citep{ma2024emotion2vec,xu2025qwen3omni}. For Qwen3-Omni, confidence over ten votes must be at least 0.90 and entropy at most 0.20. This screening yields 1,691,770 clips totaling 1,999.97 hours, with the source breakdown in Table~\ref{tab:humanspeech_sources}.

\begin{table}[!htbp]
\caption{HumanSpeech-1M after speech screening and segmentation.}
\label{tab:humanspeech_sources}
\centering
\small
\begin{tabularx}{\textwidth}{@{}lYrr@{}}
\toprule
\textbf{Language} & \textbf{Source} & \textbf{Clips} & \textbf{Hours} \\
\midrule
English & Affective media collection & 259,152 & 337.10 \\
English & GigaSpeech Audiobook & 254,871 & 303.79 \\
English & GigaSpeech YT & 130,254 & 187.28 \\
English & GigaSpeech POD & 108,657 & 156.90 \\
English & VoxMovies & 3,752 & 3.43 \\
\midrule
Chinese & WenetSpeech L & 511,155 & 542.48 \\
Chinese & Affective media collection & 407,950 & 441.05 \\
Chinese & EmotionTalk & 7,410 & 10.43 \\
Chinese & Hongloumeng & 2,950 & 8.14 \\
Chinese & CSEMOTIONS & 2,844 & 6.19 \\
Chinese & RealMAN & 1,715 & 2.34 \\
Chinese & ESD & 1,060 & 0.84 \\
\midrule
Total & & 1,691,770 & 1,999.97 \\
\bottomrule
\end{tabularx}
\end{table}

The next stages select utterances for which hearing the voice can change a useful assistant response. DeepSeek-V4-Flash first makes a high-recall pass over transcripts, and GPT-5.5 then narrows the set for precision, leaving 90K candidates. They favor substantive, self-contained utterances addressed to an assistant and remove narration, news, lectures, quoted dialogue, missing context, explicit affect statements, and requests for physical action. Gemini-3.5-Flash then listens for audible affect and response feasibility. The final branch contains approximately 75K utterances and 83 hours of speech. Table~\ref{tab:humanspeech_retained} reports the detailed counts.

\begin{table}[!htbp]
\caption{HumanSpeech-1M after dialogue filtering.}
\label{tab:humanspeech_retained}
\centering
\small
\begin{tabularx}{\textwidth}{@{}lYr@{}}
\toprule
\textbf{Language} & \textbf{Source} & \textbf{Retained} \\
\midrule
Chinese & WenetSpeech L & 26,854 \\
Chinese & Affective media collection & 19,784 \\
English & GigaSpeech Audiobook & 9,464 \\
English & GigaSpeech YT & 6,809 \\
English & Affective media collection & 6,308 \\
English & GigaSpeech POD & 4,681 \\
Mixed & Other human speech sources & 1,029 \\
\midrule
Total & & 74,929 \\
\bottomrule
\end{tabularx}
\end{table}

\paragraph{GeneralSpeech.}
\label{app:generalspeech}
GeneralSpeech adds Chinese and English tasks in mathematics, knowledge, comprehension, and instruction following. GSM8K, MATH, and NuminaMath-1.5 supply mathematical problems, CommonsenseQA and BoolQ supply choice and Boolean questions, and T\"{u}lu~3, public question answering corpora, and internal tasks extend general instruction coverage \citep{cobbe2021gsm8k,hendrycks2021math,li2024numinamath,talmor2019commonsenseqa,clark2019boolq,lambert2024tulu3}.

Before Qwen3-TTS synthesis \citep{hu2026qwen3tts}, written inputs are normalized into natural spoken questions. This step verbalizes equations, symbols, and units, removes markup, and preserves all numbers, conditions, options, logical relations, and requested answer forms. The initial 581K utterances total approximately 1,200 hours. Removing inputs that cannot be rendered as self-contained spoken questions, together with a 25-second duration limit, leaves 428K candidates. Subsequent CoT rewriting checks task validity, answer correctness, spoken response suitability, and answer type, producing approximately 270K training utterances. Table~\ref{tab:generalspeech_distribution} reports task shares before final balancing.

\begin{table}[!htbp]
\caption{GeneralSpeech task distribution before final balancing.}
\label{tab:generalspeech_distribution}
\centering
\small
\begin{tabularx}{\textwidth}{@{}Yr@{}}
\toprule
\textbf{Task family} & \textbf{Share} \\
\midrule
Explanation and comprehension & 35.8\% \\
Numeric and mathematical & 20.2\% \\
Limited generation & 20.0\% \\
Multiple choice & 13.6\% \\
Factual entity & 8.1\% \\
Boolean judgement & 2.3\% \\
\midrule
Total & 100\% \\
\bottomrule
\end{tabularx}
\end{table}

\subsection{CoT Construction}
\label{app:cot_design}

Each CoT preserves the steps needed to support its response. A step introduces evidence, derives a consequence, rules out an alternative, repairs a contradiction, or verifies a conclusion. We remove restatement, repeated paraphrases, empty checks, and reasoning after the conclusion. This gives concise supervision while retaining the derivation, following work on compact reasoning traces \citep{hassid2025overthink,wu2025moreisless,xu2025chainofdraft}.

\paragraph{Dialogue and emotion classification.}
Gemini-3.5-Flash listens to each retained waveform and generates a CoT and response using the states in Table~\ref{tab:dialogue_states}. Dialogue traces connect audible evidence to a conversational need and response plan. Emotion traces compare acoustic evidence across candidate classes and end with an affective description. Perceived age, gender, accent, and background sound enter the dialogue trace only when they affect the plan, phrased as acoustic impressions rather than identity claims. Final response wording and emotion labels appear in the response field.

\begin{table}[!htbp]
\caption{CoT states for dialogue and emotion classification. The dialogue Risk state is included when ambiguity or safety changes the response plan.}
\label{tab:dialogue_states}
\centering
\small
\begin{tabularx}{\textwidth}{@{}Yl@{}}
\toprule
\textbf{State} & \textbf{Content} \\
\midrule
\multicolumn{2}{@{}l}{\textbf{Dialogue}} \\
Cue & Audible properties that affect the reply, such as a shaky voice and long pauses \\
Need & The conversational need suggested by those cues, such as reassurance before advice \\
Risk & Ambiguity or safety considerations that affect the reply \\
Plan & The response act, order, and tone, such as brief validation followed by one next step \\
\midrule
\multicolumn{2}{@{}l}{\textbf{Emotion classification}} \\
Cue & Pitch, energy, pace, rhythm, voice quality, and pauses \\
Rule out & An acoustic distinction from the closest competing emotion \\
State & Valence, arousal, and control, with the final label reserved for the response \\
\bottomrule
\end{tabularx}
\end{table}

\paragraph{General reasoning.}
Qwen3-Omni-Thinking first generates a CoT and response \citep{xu2025qwen3omni}. Gemini-3.5-Flash applies a shared shortening rule to produce a comparison version and a verified training version. The comparison version preserves the original answer exactly, including errors, so each original and shortened trace is paired with the same question and answer. This pairing supports the Qwen2.5-Omni experiment in Section~\ref{sec:exp_analysis} \citep{xu2025qwen25omni}. Because the shared rule also removes inconsistent steps, a comparison trace can repair local slips, as in Table~\ref{tab:examples_resolution}, while its answer stays fixed. The training version additionally verifies the answer and repairs both the CoT and response when needed. It retains 95.68\% of rollouts after answer verification, regenerates 4.16\% with incorrect answers, and revises the remaining 0.16\% for safety. Verified training outputs enter the main mixture.

\paragraph{Validation.}
\label{app:validation}
Generated examples are checked for valid audio and transcript alignment, required CoT states, answer separation, and consistency between reasoning and response. CoTs must stay below 120 tokens and contain useful intermediate steps. Final answer strings, option letters, emotion labels, and dialogue response wording belong in the response alone. Persona checks reject unsupported identity or capability claims, diagnoses, and promises of physical action. A failed trace is regenerated or rejected, preserving complete reasoning rather than truncating it. Table~\ref{tab:cot_lengths} summarizes the resulting lengths.

\begin{table}[!htbp]
\caption{CoT token lengths and mean state counts after rewriting. A state is one reasoning step; the dialogue and emotion branches use the named states in Table~\ref{tab:dialogue_states}. All traces contain fewer than 120 tokens.}
\label{tab:cot_lengths}
\centering
\small
\begin{tabularx}{\textwidth}{@{}Yrrrr@{}}
\toprule
\textbf{Branch} & \textbf{p10} & \textbf{p50} & \textbf{p90} & \textbf{Mean states} \\
\midrule
LIME-440K & 50 & 71 & 95 & 3.67 \\
EmotionCoT-35K & 41 & 46 & 52 & 3.00 \\
HumanSpeech-1M & 49 & 65 & 89 & 3.31 \\
GeneralSpeech & 40 & 63 & 88 & 2.10 \\
\bottomrule
\end{tabularx}
\end{table}

\subsection{Prompt Templates}
\label{app:prompts}

The templates below summarize the teacher instructions. Dialogue and emotion generation use the state definitions in Table~\ref{tab:dialogue_states}. Each generation template requires a CoT below 120 tokens and a separate final response.

\begin{promptbox}{Transcript filtering}
\small
\textbf{High-recall pass}\quad Given a transcript with language, duration, speaker, and scene metadata, keep natural, substantive, self-contained speech from one speaker suitable for an assistant response. Retain uncertain cases with a plausible conversational interpretation. Reject narration, news, lectures, recitation, quoted dialogue, missing context, severe transcription errors, system messages, role impersonation, and requests for physical action.

\textbf{High-precision pass}\quad Review the retained utterance for direct address to an assistant and enough context for a useful reply. Keep it when the waveform could change the interpretation, tone, or content of that reply and the transcript leaves the affective condition implicit.

\textbf{Output}\quad Return \emph{Keep} or \emph{Reject} with one concise reason at each pass.
\end{promptbox}

\begin{promptbox}{Dialogue and emotion supervision}
\small
\textbf{Dialogue}\quad Listen to the waveform and check audio quality, transcript agreement, conversational delivery, assistant-directed speech, implicit affect, and response feasibility. If it passes, write Cue, Need, optional Risk, and Plan states, followed by a concise response that executes the plan. Otherwise return the decisive rejection reason.

\textbf{Emotion classification}\quad Predict emotion from the waveform using the allowed labels, independently of the source label. Write Cue, Rule out, and State, grounding the comparison in audible properties. Place the final emotion label in the response.
\end{promptbox}

\begin{promptbox}{Text normalization for speech synthesis}
\small
Rewrite the written question as natural spoken language, preserving every number, condition, option, unit, and logical relation. Expand equations, symbols, markup, and abbreviations as needed for unambiguous pronunciation. Return only the normalized question, leaving it unsolved.
\end{promptbox}

\begin{promptbox}{GeneralSpeech rewriting}
\small
\textbf{Shared instruction}\quad Given the original CoT and response, retain necessary facts, operations, and checks. Remove restatement, redundant operations, empty checks, and reasoning after the conclusion. Keep the CoT below 120 tokens and place the final answer in the response.

\textbf{Comparison version}\quad Preserve the original answer exactly, including any error, and shorten only the CoT.

\textbf{Verified training version}\quad Independently verify the answer. Retain a concise valid derivation when correct. Otherwise solve the task and rewrite both the CoT and response. Make the response natural when spoken.

\textbf{Output}\quad Return the CoT and response for the requested version.
\end{promptbox}

\subsection{Plan Execution}
\label{app:quality_evaluation}

A dialogue CoT is useful supervision only if its Plan state shapes the response rather than describing it after the fact. We therefore ask an automatic judge whether each response carries out the plan stated in its CoT, and repeat the judgment on controls that pair the same CoT with a randomly chosen response from another example. Table~\ref{tab:plan_response_consistency} reports both conditions.

\begin{table}[!htbp]
\caption{Plan execution on 300 dialogue examples. The judge rates whether the response carries out the plan stated in the CoT. Controls pair each CoT with a randomly chosen response from another example.}
\label{tab:plan_response_consistency}
\centering
\small
\begin{tabularx}{\textwidth}{@{}Yrrr@{}}
\toprule
\textbf{Pairing} & \textbf{Fully executed} & \textbf{Partly executed} & \textbf{Contradictory} \\
\midrule
True pair & 91.3\% & 8.0\% & 0.7\% \\
Mismatched control & 6.0\% & 31.0\% & 63.0\% \\
\bottomrule
\end{tabularx}
\end{table}

True pairs are judged fully executed in $91.3\%$ of cases against $6.0\%$ for the controls, so the judge separates matched from mismatched reasoning rather than accepting any response, and the plans are specific enough that an arbitrary response does not satisfy them.

\endgroup

\subsection{Illustrative Examples}
\label{app:cot_examples}

Ten selected English examples illustrate how our supervision connects acoustic evidence to a response and condenses general reasoning. All displayed CoTs and responses are complete records, with only typography, line breaks, and mathematical notation normalized. Transcripts aid readability; acoustic descriptions come from the generated CoT annotations.

\begingroup
\definecolor{exampleaccent}{RGB}{24,91,96}
\definecolor{examplefill}{RGB}{236,245,245}
\setlength{\tabcolsep}{7pt}
\renewcommand{\arraystretch}{1.12}
\newcommand{\examplecase}[2]{%
  \rowcolor{examplefill}
  \multicolumn{2}{@{}p{\textwidth}@{}}{\strut\textcolor{exampleaccent}{\textbf{#1}}\hfill{\footnotesize #2}\strut}\\[2pt]}
\newcommand{\exampleinput}[2]{%
  \multicolumn{2}{@{}p{\textwidth}@{}}{\textbf{#1:} #2}\\[2pt]}
\newcommand{\examplestate}[2]{\textbf{#1:} #2\par}

\begin{table}[H]
\caption{Empathetic dialogue examples. The response plan changes with the voice and conversational need: make room for reluctance, share a success, offer a manageable boundary, or respect a request for space.}
\label{tab:examples_dialogue}
\centering
\small
\begin{tabularx}{\textwidth}{@{}>{\hsize=1.1\hsize\linewidth=\hsize}Y>{\hsize=.9\hsize\linewidth=\hsize}Y@{}}
\toprule
\textbf{Constructed CoT} & \textbf{Response} \\
\midrule
\examplecase{D1. Good news, a heavy voice}{LIME-440K}
\exampleinput{Transcript}{``They promoted me to lead the sustainability project...''}
\examplestate{Cue}{low pitch, slow tempo, heavy sighing breath at the start, downward intonation $\to$ heavy reluctance, overwhelm}
\examplestate{Need}{validation that a promotion can feel like a burden rather than a reward}
\examplestate{Risk}{congratulating them forces a positive reaction they do not feel}
\examplestate{Plan}{acknowledge the heavy weight of the news, then ask what makes this feel daunting}
& That sounds like a massive undertaking, and from your tone, it feels more like a heavy weight than a celebration. What is the part of this new role that is weighing on you the most right now? \\
\addlinespace[3pt]
\examplecase{D2. Letting a success be a success}{LIME-440K}
\exampleinput{Transcript}{``The professor approved my thesis draft without any changes!''}
\examplestate{Cue}{bright, rapid tempo, rising pitch, audible smile $\to$ relief, pride}
\examplestate{Need}{validation of massive effort behind clean pass}
\examplestate{Risk}{focusing on next steps ruins rare moment of absolute victory}
\examplestate{Plan}{match triumphant energy, highlight zero-change approval, invite celebration}
& Without any changes? That is practically unheard of! You must be walking on air right now after all that work. How are you going to celebrate tonight? \\
\addlinespace[3pt]
\examplecase{D3. Hearing the need behind a technical question}{HumanSpeech-1M}
\exampleinput{Transcript}{``you can just turn off your phone forever is that feasible?''}
\examplestate{Cue}{high pitch, rapid tempo, sharp upward inflection on `forever' and `feasible' $\to$ exasperated overwhelm}
\examplestate{Need}{validation of the urge to escape digital noise}
\examplestate{Risk}{treating it as a literal technical query ignores the underlying burnout}
\examplestate{Plan}{acknowledge the fantasy of disappearing from the grid, then address the practical reality gently}
& It is completely understandable to want to just pull the plug and disappear from the grid for a while. While you can physically turn it off, the real challenge is dealing with the people who expect you to be reachable. Would you want to try setting up some strict boundaries first, like a do-not-disturb schedule? \\
\addlinespace[3pt]
\examplecase{D4. Knowing when to stop asking}{HumanSpeech-1M}
\exampleinput{Transcript}{``i ain't got nothing to say to nobody.''}
\examplestate{Cue}{low pitch, heavy vocal tension, sharp downward slide on the final word $\to$ defensive anger, high intensity}
\examplestate{Need}{to draw a hard boundary and have it respected without pushback}
\examplestate{Risk}{asking why they are shut down will feel like an interrogation and provoke more anger}
\examplestate{Plan}{acknowledge the boundary immediately, offer a quiet exit}
& Understood. I will step back and leave you be. If you change your mind later, I'll be right here. \\
\bottomrule
\end{tabularx}
\end{table}

\begin{table}[H]
\caption{Speech emotion recognition (SER) examples from three human speech sources. The task is to identify emotion from the audio. Each CoT connects audible evidence to a competing-class exclusion and an affective state; the selected label appears only in the response.}
\label{tab:examples_ser}
\centering
\small
\begin{tabularx}{\textwidth}{@{}>{\hsize=1.65\hsize\linewidth=\hsize}Y>{\hsize=.35\hsize\linewidth=\hsize}Y@{}}
\toprule
\textbf{Constructed CoT} & \textbf{Response} \\
\midrule
\examplecase{S1. Amusement in a factual utterance}{Expresso}
\exampleinput{Transcript}{``There's an alarm set for today at two thirty AM, another for tomorrow at two fifteen PM, and for Wednesday at seven PM.''}
\examplestate{Cue}{breathy giggling, wide pitch variation, and bright, smiling vocal resonance.}
\examplestate{Rule out}{confused --- presence of bubbling laughter and rising, playful pitch contours.}
\examplestate{State}{high-arousal positive valence with spontaneous amusement.}
& \textbf{happy} \\
\addlinespace[7pt]
\examplecase{S2. Strained restraint versus overt anger}{IEMOCAP}
\exampleinput{Transcript}{``Okay, no, look, I- I- I- I'm really -- I'm trying to be patient. I know this isn't your fault. I know that you didn't do it''}
\examplestate{Cue}{rapid and uneven tempo, audible sighs, tense vocal quality, frequent self-interruptions}
\examplestate{Rule out}{angry --- lacks the high-energy acoustic projection and sharp volume spikes}
\examplestate{State}{moderate-to-high arousal negative valence, showing signs of strained self-control}
& \textbf{frustrated} \\
\addlinespace[7pt]
\examplecase{S3. A sudden realization versus sustained joy}{MSP-Podcast}
\exampleinput{Transcript}{``and then you look up and you go, `wait a minute. whoo, things are changing.'\,''}
\examplestate{Cue}{sudden pitch rise, wide dynamic range, breathy exclamation, rapid tempo}
\examplestate{Rule out}{happy --- lacks sustained positive resonance, dominated by sudden pitch shifts}
\examplestate{State}{high arousal, sudden cognitive shift, startled realization}
& \textbf{surprise} \\
\bottomrule
\end{tabularx}
\end{table}

\begin{table}[H]
\caption{General reasoning: the division identity and remainder bound determine the smallest valid divisor, making repeated checks unnecessary. CoT: $529\to62$ tokens ($8.5\times$ shorter).}
\label{tab:examples_division}
\centering
\small
\begin{tabularx}{\textwidth}{@{}>{\hsize=1.16\hsize\linewidth=\hsize}Y>{\hsize=.84\hsize\linewidth=\hsize}Y@{}}
\toprule
\examplecase{G1. Minimum feasible dividend}{GeneralSpeech}
\exampleinput{Question}{When two numbers are divided, the quotient is 12 and the remainder is 3. What is the smallest possible dividend?}
\textbf{Original CoT: Qwen3-Omni-Thinking} & \textbf{Our rewritten CoT} \\
\midrule
Okay, let's see. The problem says when two numbers are divided, the quotient is 12 and the remainder is 3. We need to find the smallest possible dividend. Hmm, right. So first, I remember that when you divide numbers, there's a relationship between the dividend, divisor, quotient, and remainder. Let me recall the formula: Dividend = Divisor $\times$ Quotient + Remainder. Yeah, that's right. So if we let the divisor be $d$, the quotient is 12, remainder is 3, then the dividend $N$ is $N=12d+3$.
\par\smallskip
But wait, the remainder has to be less than the divisor, right? Because if the remainder was equal to or larger than the divisor, you could divide again. So the remainder $r$ must satisfy $0\leq r<d$. In this case, the remainder is 3, so $3<d$. That means the divisor $d$ has to be greater than 3. So the smallest possible divisor would be 4, since it has to be an integer greater than 3. Let me check that. If $d=4$, then the dividend would be $12\times4+3=48+3=51$. Let's verify: 51 divided by 4 is 12 with a remainder of 3, because 4 times 12 is 48, and 51 minus 48 is 3. Yep, that works.
\par\smallskip
Is there a smaller divisor? Well, if $d=3$, then the remainder would have to be less than 3, but the remainder here is 3, which is not less than 3, so that's not allowed. If $d=2$, same problem, remainder 3 is bigger than 2, so you could divide more. So the smallest possible divisor is 4, which gives the smallest dividend of 51. Let me just make sure. Let's try $d=4$: $51\div4=12.75$, but in integer division, it's 12 with remainder 3. Correct. If we take $d=5$, then dividend is $12\times5+3=63$, which is bigger, so 51 is smaller. So yeah, 51 is the smallest possible dividend.
&
\examplestate{Given}{Dividend = Divisor $\times12+3$, and remainder 3 must be less than divisor.}
\smallskip
\examplestate{Step}{The smallest integer divisor satisfying $3<d$ is 4.}
\smallskip
\examplestate{Step}{Dividend $=12\times4+3=48+3$.}
\par\medskip
\textcolor{exampleaccent}{\textbf{Response}}\par\smallskip
The smallest possible dividend is 51, because the divisor must be at least 4, and 12 times 4 plus 3 equals 51. \\
\bottomrule
\end{tabularx}
\end{table}

\begin{table}[H]
\caption{General reasoning: establish consecutive angles, derive their common scale, and select the requested angle. CoT: $517\to86$ tokens ($6.0\times$ shorter).}
\label{tab:examples_angles}
\centering
\small
\begin{tabularx}{\textwidth}{@{}>{\hsize=1.16\hsize\linewidth=\hsize}Y>{\hsize=.84\hsize\linewidth=\hsize}Y@{}}
\toprule
\examplecase{G2. Selecting the obtuse angle}{GeneralSpeech}
\exampleinput{Question}{The ratio of two interior angles of a parallelogram is $4:5$. What is the degree measure of one of the two obtuse angles of the parallelogram?}
\textbf{Original CoT: Qwen3-Omni-Thinking} & \textbf{Our rewritten CoT} \\
\midrule
Okay, let's try to figure out this problem. So, we have a parallelogram, and the ratio of two interior angles is $4:5$. We need to find the measure of one of the obtuse angles. Hmm, first, I remember that in a parallelogram, opposite angles are equal, and consecutive angles are supplementary. That means if one angle is, say, $x$, the next one is $180-x$, right? Because they add up to 180 degrees.
\par\smallskip
So, the problem says the ratio of two interior angles is $4:5$. Let's assume these are consecutive angles because if they were opposite angles, they would be equal, so their ratio would be $1:1$, which isn't $4:5$. So, they must be consecutive angles. Let me confirm that. In a parallelogram, adjacent angles are supplementary, so they add up to 180 degrees. So, if the ratio is $4:5$, let's let the measures be $4k$ and $5k$. Then, $4k+5k=180$ degrees. Because they are consecutive angles, they should add up to 180. Let me check that.
\par\smallskip
Yes, in a parallelogram, consecutive angles are supplementary. So, $4k+5k=180$. That would be $9k=180$, so $k=20$. Then, the angles would be $4\times20=80$ degrees and $5\times20=100$ degrees. Now, which one is obtuse? Obtuse angles are greater than 90 degrees, so 100 degrees is obtuse. So, the measure of one of the obtuse angles is 100 degrees. Let me just make sure.
\par\smallskip
In a parallelogram, opposite angles are equal. So, if two consecutive angles are 80 and 100, then the other two angles are also 80 and 100. So, there are two acute angles (80 degrees) and two obtuse angles (100 degrees). So, the obtuse angles are 100 degrees each. That makes sense. Let me verify the ratio: $80:100$ simplifies to $4:5$, which matches the problem statement. Yep, that works. So, the answer should be 100 degrees.
&
\examplestate{Given}{parallelogram interior angles ratio $4:5$, opposite angles equal, consecutive supplementary.}
\smallskip
\examplestate{Step}{The ratio must be for consecutive angles. Let them be $4k$ and $5k$. Then $4k+5k=180\Rightarrow9k=180\Rightarrow k=20$.}
\smallskip
\examplestate{Step}{The two angles are $4k$ and $5k$. The larger is obtuse.}
\par\medskip
\textcolor{exampleaccent}{\textbf{Response}}\par\smallskip
The obtuse angle is 100 degrees. \\
\bottomrule
\end{tabularx}
\end{table}

\begin{table}[H]
\caption{General reasoning: explaining an observation from a physical law. The rewrite keeps the diffraction limit and the required separation interval. It also resolves the original CoT's reversed inequalities in its first ``Therefore'' paragraph; the original later gives the correct interval verbally. CoT length: $385\to66$ tokens ($5.8\times$ shorter).}
\label{tab:examples_resolution}
\centering
\small
\begin{tabularx}{\textwidth}{@{}>{\hsize=1.16\hsize\linewidth=\hsize}Y>{\hsize=.84\hsize\linewidth=\hsize}Y@{}}
\toprule
\examplecase{G3. Resolving two stars in blue light}{GeneralSpeech}
\exampleinput{Question}{Why might some stars appear double in blue light, though they could not be resolved in red light with the same telescope?}
\textbf{Original CoT: Qwen3-Omni-Thinking} & \textbf{Our rewritten CoT} \\
\midrule
We are considering the phenomenon where a star appears double in blue light but not in red light with the same telescope.
\par\smallskip
The key point is the diffraction limit of the telescope. The angular resolution ($\theta$) of a telescope is given by the Rayleigh criterion:
\par $\theta=1.22\times(\lambda/D)$
\par where:
\par $\lambda$ = wavelength of light
\par $D$ = diameter of the telescope's aperture
\par\smallskip
Since blue light has a shorter wavelength ($\lambda_{\mathrm{blue}}$) than red light ($\lambda_{\mathrm{red}}$), the angular resolution for blue light is better (smaller $\theta_{\mathrm{blue}}$) than for red light ($\theta_{\mathrm{red}}$).
\par\smallskip
Therefore, if two stars are separated by an angular distance that is greater than $\theta_{\mathrm{red}}$ (so they are not resolved in red light) but less than $\theta_{\mathrm{blue}}$ (so they are resolved in blue light), then we would see them as double in blue light but not in red.
\par\smallskip
However, note that the Rayleigh criterion is an approximation and the actual resolution also depends on the point spread function and other factors. But the main idea is that the resolving power of a telescope improves as the wavelength decreases.
\par\smallskip
Example:
\par Let $\lambda_{\mathrm{red}}\approx700$ nm, $\lambda_{\mathrm{blue}}\approx400$ nm.
\par Then $\theta_{\mathrm{blue}}/\theta_{\mathrm{red}}=\lambda_{\mathrm{blue}}/\lambda_{\mathrm{red}}\approx400/700\approx0.57$.
\par\smallskip
So the resolving power in blue is about 1.75 times better than in red.
\par\smallskip
Therefore, if two stars are separated by an angle that is between $\theta_{\mathrm{blue}}$ and $\theta_{\mathrm{red}}$, they will be resolved in blue but not in red.
\par\smallskip
Hence, the answer is: because blue light has a shorter wavelength, which allows the telescope to resolve finer details (smaller angular separations) than red light.
\par\smallskip
So the phenomenon occurs due to the wavelength dependence of the diffraction limit.
&
\examplestate{Given}{telescope angular resolution $\theta\approx1.22\lambda/D$}
\smallskip
\examplestate{Step}{blue light has shorter $\lambda$ than red light $\to$ smaller $\theta$ for blue}
\smallskip
\examplestate{Step}{if two stars have angular separation $\Delta\theta$ such that $\theta_{\mathrm{blue}}<\Delta\theta<\theta_{\mathrm{red}}$, they are unresolved in red but resolved in blue}
\par\medskip
\textcolor{exampleaccent}{\textbf{Response}}\par\smallskip
Stars appear double in blue light because the telescope's diffraction limit depends on wavelength. Blue light, with its shorter wavelength, gives better angular resolution than red light. If the stars' separation falls between the red and blue resolution limits, they blend into one in red but appear as two in blue. \\
\bottomrule
\end{tabularx}
\end{table}
\clearpage
\endgroup

\end{document}